\documentclass{article}

\usepackage{microtype}
\usepackage{graphicx}
\usepackage{subcaption}
\usepackage{booktabs} %
\usepackage[T1]{fontenc}
\usepackage{hyperref}

\usepackage{amssymb}
\usepackage{tikz}
\usepackage{enumitem}
\usetikzlibrary{arrows.meta, positioning, fit, backgrounds, calc}
\usepackage{url}
\usepackage{color}
\usepackage{pdflscape}  %
\usepackage{array} 
\usepackage{enumitem}
\usepackage{longtable}
\usepackage{listings}
\usepackage{xcolor}
\usepackage[dvipsnames]{xcolor}
\usepackage{hyperref}

\usepackage[textsize=tiny,textwidth=1in]{todonotes}

\usepackage[preprint]{icml2026}

\usepackage{amsmath}
\usepackage{amssymb}
\usepackage{mathtools}
\usepackage{amsthm}
\usepackage{tabularx}

\usepackage[capitalize,noabbrev]{cleveref}

\theoremstyle{plain}

\theoremstyle{definition}

\theoremstyle{remark}

\usepackage{caption}                                                                       
\usepackage[textsize=tiny]{todonotes}

\icmltitlerunning{ML-Tool-Bench: Tool-Augmented Planning for ML Tasks}

\DeclareMathOperator*{\argmax}{arg\,max}
\begin{document}

\twocolumn[
  \icmltitle{ML-Tool-Bench: Tool-Augmented Planning for ML Tasks}

  \icmlsetsymbol{equal}{*}

  \begin{icmlauthorlist}
    \icmlauthor{Yaswanth Chittepu}{equal,yyy}
    \icmlauthor{Raghavendra Addanki}{comp}
    \icmlauthor{Tung Mai}{comp}
    \icmlauthor{Anup Rao}{comp}
    \icmlauthor{Branislav Kveton}{comp}
  \end{icmlauthorlist}

  \icmlaffiliation{yyy}{University of Massachusetts Amherst}
  \icmlaffiliation{comp}{Adobe Research}

  \icmlcorrespondingauthor{Yaswanth Chittepu}{ychittepu@umass.edu}

  \icmlkeywords{Machine Learning, ICML}

  \vskip 0.3in
]

\printAffiliationsAndNotice{}  %

\begin{abstract}

     End-to-end machine learning requires coordinating long sequences of interdependent operations: data loading, preprocessing, feature engineering, model training, and prediction. While LLM agents have shown promise through direct code generation, such approaches are brittle and tightly couple reasoning with execution. Tool-augmented agents offer an alternative with modular building blocks that decouple planning from code, yet existing benchmarks evaluate only shallow trajectories (1-3 tool calls). Real ML workflows demand 20+ sequential invocations producing complex artifacts like dataframes and fitted models that must persist across the trajectory. We introduce ML-Tool-Bench, the first benchmark for long-horizon, tool-augmented planning on ML tasks, comprising 61 specialized tools and 15 Kaggle challenges. We show that ReAct and LATS fail to produce valid tool sequences for these complex pipelines. To address this, we propose shaped rewards with stage-wise feedback, hierarchical task decomposition with tool masking, and in-memory named object management for storing and retrieving intermediate results. With GPT-4o, our methods improve over ReAct by 16.52 percentile positions (median across challenges), demonstrating that structured guidance is critical for effective tool-augmented planning.
Code and benchmark: \url{https://github.com/yaswanthchittepu/ML-Tool-Bench}.
   \vspace*{-3ex}
\end{abstract}

\section{Introduction}
\label{sec: introduction}
    
  Solving an end-to-end machine learning (ML) task requires coordinating a sequence of interdependent operations: loading data, handling missing values, encoding categorical features,     
  engineering new features, fitting models, and generating predictions. Autonomous agents capable of executing such pipelines represent an active area of research                          
  \citep{grosnit2024largelanguagemodelsorchestrating, toledo2025airesearchagentsmachine, Yang2025ReinforcementLF, chan2025mlebenchevaluatingmachinelearning}. Such agents must manage       
  intermediate results and adapt their strategies based on evolving context. Large language models (LLMs) have recently been explored as the foundation for such agents                     
  \citep{grosnit2024largelanguagemodelsorchestrating, chan2025mlebenchevaluatingmachinelearning, huang2024mlagentbenchevaluatinglanguageagents}. Early work has primarily focused on direct 
  code generation, where the agent produces Python code for completing a given ML task \citep{grosnit2024largelanguagemodelsorchestrating, chan2025mlebenchevaluatingmachinelearning,       
  huang2024mlagentbenchevaluatinglanguageagents, toledo2025airesearchagentsmachine}. This paradigm has shown promise on competitive benchmarks inspired by Kaggle challenges, with some     
  approaches achieving performance comparable to a Kaggle Master \citep{grosnit2024largelanguagemodelsorchestrating, chan2025mlebenchevaluatingmachinelearning}. Several benchmarks have    
  been proposed to evaluate LLM performance on such tasks \citep{chan2025mlebenchevaluatingmachinelearning, huang2024mlagentbenchevaluatinglanguageagents,                                  
  qiang2025mledojointeractiveenvironmentsempowering, jing2025dsbenchfardatascience, zhang2025datascibenchllmagentbenchmark}. However, direct code generation is prone to key weaknesses:    
  generated code is brittle \citep{abbassi2025taxonomyinefficienciesllmgeneratedpython, liu2025codemindevaluatinglargelanguage}, debugging typically requires multiple iterations, and      
  reasoning is tightly coupled with execution \citep{liu2025codemindevaluatinglargelanguage, chen2025steeringlargelanguagemodels}.

  An alternative paradigm equips LLMs with external tools, yielding tool-augmented agents that decide which tools to invoke and in what sequence. Tools offer modular, reusable building    
  blocks for data-science workflows: from preprocessing to training and evaluation. This design has proven effective in domains including web navigation                                    
  \citep{zhou2024webarenarealisticwebenvironment}, operating systems \citep{bonatti2024windowsagentarenaevaluating}, and code interpretation                                                
  \citep{huang2024mlagentbenchevaluatinglanguageagents}, yet its potential for ML workflows remains underexplored. Crucially, tool augmentation reformulates the problem as planning in a   
  large action space: the agent must coordinate multi-step trajectories and retrieve and reuse intermediate artifacts. Because the agent is restricted to a curated toolset, tool-augmented 
  approaches decouple high-level reasoning from low-level code execution, improving modularity, reliability, and safety.                                                                    
\begin{table*}[!ht]   
\footnotesize           
  \centering                                                                                                                                                                                
  \caption{Comparison of representative benchmarks for evaluating tool-augmented LLM Agents.}                                                                                               
  \label{tab:benchmark_comparison}                                                                                                                                                          
  \begin{tabular}{p{3.8cm} p{3.6cm} p{2.0cm} p{2.4cm} p{3.0cm}}                                                                                                                             
  \toprule                                                                                                                                                                                  
  \textbf{Benchmark} & \textbf{Focus} & \textbf{Tool Calls} & \textbf{Planning Depth} & \textbf{Artifact Management} \\                                                                     
  \midrule                                                                                                                                                                                  
  ML-Tool-Bench (Ours) & Multi-step planning & 20+ & Deep & Complex objects \\                                                                                                              
  ToolBench \citep{xu2023toolmanipulationcapabilityopensource} & Tool selection & 1--3 & Shallow & String-based \\                                                                          
  BFCL \citep{patil2025bfcl} & Argument extraction & 1--3 & Shallow & Primitive types \\                                                                                                    
  MLE-bench \citep{chan2025mlebenchevaluatingmachinelearning} & Code generation & N/A & N/A & File system \\                                                                                
  MLAgent-bench \citep{huang2024mlagentbenchevaluatinglanguageagents} & Code generation & N/A & N/A & File system \\                                                                        
  \bottomrule                                                                                                                                                                               
  \end{tabular}                                                                                                                                    \vspace*{-4ex}                                         
  \end{table*}                                                                                                                                                                              
                                                                                                                                                                                                            
  Existing benchmarks for tool use fall short on long-horizon planning. Most benchmarks evaluate whether agents can select correct tools and valid arguments. The Berkeley Function-Calling 
  Leaderboard (BFCL) \citep{patil2025bfcl} measures single, parallel, and multiple function calling, and BFCL-v3 extends this to multi-turn settings. However, even BFCL-v3 evaluates       
  trajectories of only 1 to 3 tool calls with primitive-typed arguments (strings, integers, floats). ML workflows, by contrast, require 20 or more sequential tool invocations, producing   
  intermediate artifacts (dataframes, fitted models) that must be persisted and reused across the trajectory. Similarly, ToolBench \citep{xu2023toolmanipulationcapabilityopensource} spans 
  single-step and multi-step action generation but restricts inputs to string-based arguments, limiting its ability to test artifact management at scale.

  In this work, we introduce ML-Tool-Bench, the first benchmark to evaluate the long-horizon planning capabilities of tool-augmented LLMs, using \emph{tabular} machine learning challenges from Kaggle as our testbed. We provide a curated suite of 61 tools and assess performance across 15 Kaggle challenges spanning regression and classification. To handle intermediate artifacts (dataframes, fitted models) that arise during multi-step workflows, we adopt a \emph{scratchpad-augmented} scheme: tools accept references to named objects, and agents assign names to tool outputs. This design supports arbitrarily large artifacts and enables reversible branching during tree search. A comparison of benchmarks is provided in Table~\ref{tab:benchmark_comparison}.

We evaluate multiple planning algorithms on ML-Tool-Bench. Simple methods like ReAct \citep{yao2023reactsynergizingreasoningacting}, which plan unidirectionally, struggle to produce     
  valid trajectories and cannot recover from early mistakes. Monte Carlo Tree Search methods \citep{Kocsis2006BanditBM, silver2016mastering} such as LATS                                   
  \citep{zhou2024languageagenttreesearch}, which use LLMs as value estimators, also underperform: as trajectories lengthen and artifacts accumulate in context, the evaluator LLM provides  
  inconsistent scores, causing the search to favor failed branches over successful ones. To address these limitations, we propose two approaches: (1) \textbf{Shaped Rewards}, combining    
  deterministic, stage-wise rewards with targeted textual feedback, and (2) \textbf{Task Decomposition}, which decomposes the problem into a sequence of subtasks with tool masking. Both   
  approaches outperform the baselines, yielding more valid and higher-performing trajectories.                                                                                              
                                                                                                                                                                                            
  Our contributions are:                                                                                                                                                                    
  \begin{enumerate}[leftmargin=*, nosep]                                                                                                                                                    
  \item \textbf{ML-Tool-Bench}, the first benchmark for long-horizon, tool-augmented planning on end-to-end ML workflows (61 tools, 15 Kaggle challenges).                                  
  \item \textbf{Scratchpad-augmented planning}, a named-object management scheme that supports arbitrarily large artifacts and reversible branching during tree search.                     
  \item \textbf{MCTS-based planning methods}: \emph{MCTS-Shaped}, which replaces unreliable LLM value estimates with deterministic, stage-wise rewards; and \emph{Hierarchical MCTS},   
  which decomposes problems into subtasks with tool masking. On GPT-4o (median across challenges), Hierarchical MCTS improves over LATS by 9.93 percentile points and over ReAct by 16.52   
  percentile points, establishing strong improvements for tool-augmented ML planning.                                                                                                                  
  \end{enumerate}

\section{Related Work}
\label{sec: relatedwork}

  \textbf{Machine Learning Benchmarks for AI Agents.}                                                                                                                                       
  Existing ML benchmarks primarily evaluate agents that generate code directly. MLE-bench \citep{chan2025mlebenchevaluatingmachinelearning} curates 75 Kaggle challenges and reports that   
  OpenAI's o1-preview with AIDE scaffolding \citep{jiang2025aideaidrivenexplorationspace} achieves bronze-level performance on 16.9\% of tasks. AIRA-dojo                                   
  \citep{toledo2025airesearchagentsmachine} extends this by replacing AIDE with alternative operators and MCTS \citep{Kocsis2006BanditBM}, improving medal rates from 39.6\% to 47.7\%.     
  MLAgentBench \citep{huang2024mlagentbenchevaluatinglanguageagents} provides 13 tasks where agents read/write files and execute code, achieving 37.5\% success with a ReAct-based agent    
  \citep{yao2023reactsynergizingreasoningacting}. MLE-Dojo \citep{qiang2025mledojointeractiveenvironmentsempowering} offers an interactive gym-style environment over 200+ Kaggle           
  challenges, and DS-bench \citep{jing2025dsbenchfardatascience} evaluates agents on 466 analysis and 74 modeling tasks. These benchmarks focus on code generation rather than              
  tool-augmented planning.                                                                                                                                                                  
                                                                                                                                                                                            
  \textbf{Tool Use in LLMs.}                                                                                                                                                                
  Several approaches couple reasoning with tool invocation via reinforcement learning, including ARTIST \citep{singh2025agenticreasoningtoolintegration}, ReTooL \citep{Feng2025ReToolRL},  
  StepTool \citep{Yu2024StepToolEM}, ToRL \citep{Li2025ToRLST}, and ToolPlanner \citep{Wu2024ToolPlannerAT}. Others fine-tune LLMs on tool-use traces \citep{Schick2023ToolformerLM,        
  Qin2023ToolLLMFL, Gou2023ToRAAT, patil2023gorillalargelanguagemodel}. Tree-search methods have also been applied: \citet{Zhuang2023ToolChainEA} use A* search,                            
  \citet{hao2023reasoninglanguagemodelplanning} adopt MCTS with the LLM as world model, LATS \citep{zhou2024languageagenttreesearch} combines MCTS with LLM-based value estimation, and     
  \citet{Yao2023TreeOT} explore BFS and DFS. However, these methods rely on heuristic cost functions or LLM feedback and target shallow-depth problems. TS-LLM                              
  \citep{feng2024alphazeroliketreesearchguidelarge} scales to depth 64 using learned value functions, and ReST-MCTS \citep{zhang2024restmctsllmselftrainingprocess} infers per-step rewards 
  rather than TD-$\lambda$ \citep{sutton1988td}. Among these, only LATS and Toolchain* explicitly address planning with tools. Benchmarks for tool use emphasize selection and argument     
  specification: ToolBench \citep{xu2023toolmanipulationcapabilityopensource} covers single- and multi-step tasks but underplays long-horizon coordination; BFCL \citep{patil2025bfcl}      
  evaluates function calling with shallow plans; and $\tau$-Bench \citep{yao2024taubenchbenchmarktoolagentuserinteraction} targets human-agent interaction rather than proactive planning.  
  Our work bridges ML benchmarks (code-centric) and tool-use benchmarks (shallow-depth) by evaluating long-horizon, tool-augmented planning.

\section{ML-Tool-Bench}
\label{sec: method}

Each task in ML-Tool-Bench, can be formalized as a Markov Decision Process (MDP) $(\mathcal{S}, \mathcal{A}, \mathcal{T},R)$ \citep{puterman2014markov}
The \emph{state space} $\mathcal{S}$ consists of the entire interaction history: all AI, Human, and Tool messages together with artifacts such as dataframes and ML models. Whenever a tool is executed, its observations (e.g., outputs, errors, logs) are appended to the history and folded into the state, so that {the state} maintains an up-to-date record of both conversational and artifact changes. The initial state $s_{0}$ comprises the Kaggle challenge description along with the dataset.

The \emph{action space} is defined as: $
\mathcal{A} \;=\; \big(\mathcal{A}_{\text{tool}} \cup \{\varnothing\}\big) \times \big(\mathcal{A}_{\text{reason}} \cup \{\varnothing\}\big),$
where $\mathcal{A}_{\text{tool}}$ denotes the set of all tool invocations together with their full parameterizations (not just tool identity, but also argument values and hyperparameters). This makes the benchmark challenging, since the effective size of $\mathcal{A}_{\text{tool}}$ can be very large rather than a small, discrete set. The set $\mathcal{A}_{\text{reason}}$ is the space of free-form reasoning steps, which we model as natural-language strings. The null element $\varnothing$ denotes “no action” in that component, allowing tool-only, reason-only, both, or neither at a step. Reasoning actions organize information, plan future steps, and inject prior knowledge; tool actions modify or analyze data and train/evaluate models, thereby updating the state’s artifacts. 

The \emph{transition function} $\mathcal{T}:\mathcal{S}\times\mathcal{A}\rightarrow\mathcal{S}$ maps a state–action pair $(s,a)$ to the next state by appending the messages generated by the agent’s action, appending tool messages (i.e., the observations produced), and updating artifacts accordingly. 

The \emph{reward function} $R$ evaluates progress and can be instantiated in several ways: (i) an outcome reward granted upon successful challenge completion; (ii) a shaped reward providing intermediate credit for measurable progress; or (iii) an LLM-based evaluation of the current state, using the LLM as a judge \citep{zheng2023judgingllmasajudgemtbenchchatbot}. 

\subsection{Scratchpad}

Solving an ML challenge often involves storing large dataframes, models, and other complex artifacts as they cannot be directly passed as tool inputs by an LLM. A naive workaround is to maintain a single dataframe and a model object that the agent incrementally modifies via tool calls. However, a single erroneous call can corrupt these objects, forcing a restart of the trajectory, and the agent becomes inflexible to create and reuse intermediate variables.

\begin{figure}[H]
  \centering
  \includegraphics[width=\linewidth]{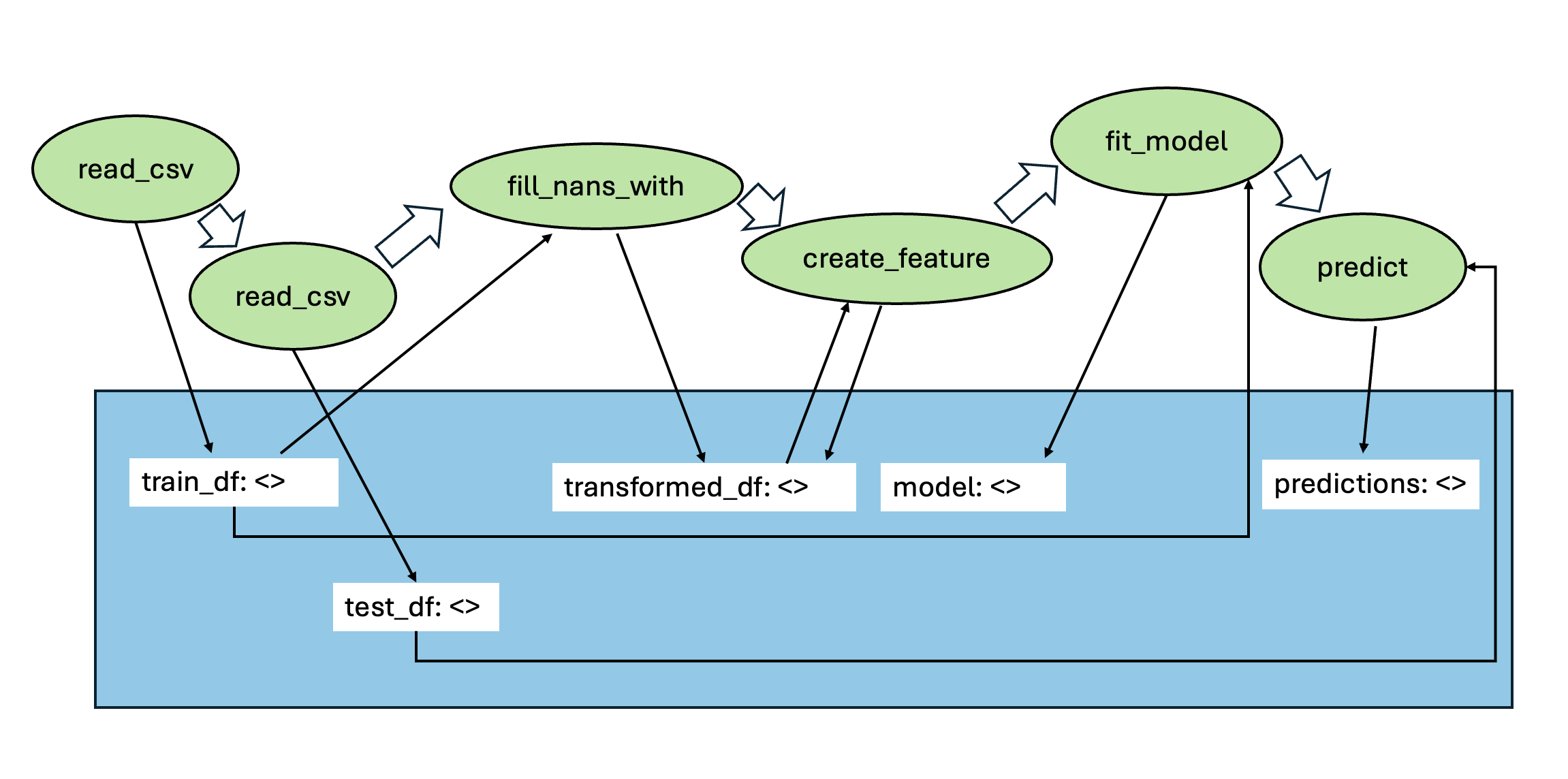} %
  \caption{An illustration of our named-object management scheme. Green circles denote tool calls; the blue rectangle denotes the scratchpad (key–value store). Arrows into tools indicate inputs; arrows out indicate outputs. Tools vary by read-write behavior: \protect\texttt{read\_csv} (set), \protect\texttt{fill\_nans\_with}/\protect\texttt{fit\_model}/\protect\texttt{predict} (get–set), and \protect\texttt{create\_feature} (override). Two \protect\texttt{read\_csv} calls load train and test data separately.}
  \label{fig:scratchpad}
  \vspace{-2ex}
\end{figure}

To address this, we adopt an in-memory, named-object management scheme: an agent assigns names to tool outputs, and tools accept references to named objects as inputs. Thus, agents can pass complex objects to tools by specifying the name under which the object is stored in the scratchpad. An illustration of this approach is presented in Figure $\ref{fig:scratchpad}$. Implementing this requires modifying tools to operate on named references rather than raw objects.

\subsection{Tools}

We grant the agent access to a curated suite of 61 tools spanning data loading, data cleaning, feature engineering, modeling, and evaluation. These tools are designed to be reasonably sufficient for solving tabular regression and classification tasks. Agent performance depends on the available toolset: in principle, a very large collection would maximize flexibility, but it results in an increased action space and complicates planning. We therefore adopt a fixed, compact tool set that trades some flexibility for a more tractable planning, while remaining adequate to solve the Kaggle challenges considered. For modeling, we restrict to tree-based learners: Random Forest, XGBoost, LightGBM, and CatBoost, and linear/logistic regression, in light of the strong performance of tree-based methods on tabular Kaggle challenges \citep{grinsztajn2022treebasedmodelsoutperformdeep}. For more information on tools and how arbitrary user defined tools are modified to operate on named references rather than objects, refer to Appendix $\ref{ap:tools}$

\subsection{Kaggle Challenges}

We select 15 tabular Kaggle ML challenges for ML-Tool-Bench: eight classification (binary and multiclass) and seven regression tasks. These tasks are chosen so that they are solvable with our tool set. Several datasets are large (e.g., New York City Taxi Fare Prediction is $\sim$2.5~GB), so we randomly sample 10{,}000 data points from each competition’s training set to keep planning computationally tractable. Because Kaggle test labels are hidden, we create an internal evaluation split by reserving 20\% of the sampled training data as a test set with ground-truth labels. We evaluate using each competition’s official metric and report agent performance as the corresponding public-leaderboard percentile. Our evaluation metric is chosen to accommodate a collection of regression and multi-class classification tasks. Note that Kaggle leaderboards are computed on a test set, the labels to which we do not have access to; our reported results are computed on our held-out test split. Later, in the results, we show that our benchmark performance serves as a reasonable proxy for the true Kaggle public leaderboard performance. For more information on the Kaggle challenges, refer to Appendix $\ref{ap:kaggle}$.

We restricted the benchmark to 15 challenges due to cost constraints ($\$5000$ for running our entire suite of experiments), which aligns with similar benchmarks like DS-Agent \citep{guo2024dsagentautomateddatascience} that used around 20 challenges. Our 15-challenge scale also aligns with other recent benchmarks involving expensive evaluation such as MLAgent-bench \citep{huang2024mlagentbenchevaluatinglanguageagents} (13 challenges) and MLE-bench \citep{chan2025mlebenchevaluatingmachinelearning} (75 challenges, but single-pass code generation not iterative planning with tree search). Benchmarks with larger scales (e.g., HumanEval \citep{chen2021evaluatinglargelanguagemodels}: 164 problems) evaluate inexpensive operations (single code generation pass) rather than multi-step planning requiring hundreds of LLM calls per trajectory

\section{Approaches}
\label{sec: methods}
\subsection{ReAct}
ReAct \citep{yao2023reactsynergizingreasoningacting} is a prompting framework that interleaves natural-language reasoning (\emph{Thought}) with tool interaction (\emph{Action}) and the subsequent \emph{Observation} from the environment due to tool calling. ReAct augments the agent's action space to include the space of language, to account for thoughts or reasoning traces that do not affect the environment. Thoughts compose useful information from the current context and update the context to support future reasoning or actions. By explicitly exposing intermediate chain-of-thought alongside tool calls, ReAct enables agents to plan, invoke tools, and revise plans based on feedback. However, ReAct is unidirectional and can neglect potential alternative continuations from certain states, leading to locally optimal solutions \citep{Zhuang2023ToolChainEA, zhou2024languageagenttreesearch}.

\subsection{Monte Carlo Tree Search (MCTS)}
MCTS \citep{Kocsis2006BanditBM}  is a search algorithm that has achieved remarkable success in challenging domains such as Go \citep{silver2016mastering} and Atari \citep{ye2021mastering}. MCTS builds a search tree where nodes correspond to states and edges correspond to actions. It comprises four phases: \emph{selection}, \emph{expansion}, \emph{simulation/rollout}, and \emph{backpropagation}. A common \emph{selection} policy uses UCT (Upper Confidence Bound for Trees) \citep{Kocsis2006BanditBM}, choosing a child $s$ of parent $p$ such that: 
    $s \in \argmax_{s \in \mathbb{C}(p)} \; V(s) \;+\; w\,\sqrt{{\ln N(p)}/{N(s)}},$

where $V(s)$ is the empirical value function, denoting the expected cumulative reward from state $s$, $N(p)$ is the parent’s visit count, $N(s)$ is the child’s visit count, $w>0$ controls exploration, and $\mathbb{C}(p)$ denotes the set of children of $p$. Upon reaching a leaf node, it is \emph{expanded} by selecting an action and adding the resulting next state as a child. From the newly expanded node, a \emph{simulation} is run until the end of the episode or a fixed depth to obtain a reward $r$, which is then \emph{backpropagated} along the trajectory to update values of all states along that trajectory:
    $V(s) \leftarrow {\left(V(s)\,(N(s)-1) + r\right)}/{N(s)}.$
MCTS is well-suited to large, irregular action spaces and provides a principled trade-off between exploration and exploitation. A pictorial illustration of MCTS is provided in Appendix $\ref{ap:approaches}$.

\subsection{Language Agent Tree Search (LATS)}
LATS \citep{zhou2024languageagenttreesearch} adapts MCTS to language agents by using LLMs both to propose actions (reasoning steps or tool calls) and to evaluate node values. At each expansion, the policy LLM suggests candidates, and an evaluator LLM scores partial trajectories based on estimated progress toward the task objective. The value of a state is taken to be a weighted average of the evaluator LLM’s score and a self-consistency score \citep{wang2022self}, which upweights frequent candidates in the expansion stage. 

LATS has shown improvements over purely reactive methods, such as ReAct \citep{yao2023reactsynergizingreasoningacting} on complex tasks. However, its value estimates can be noisy, and the effective planning depth may be limited by inconsistencies in evaluator scoring.

\subsection{MCTS-Shaped}
\label{subsec:mcts-shaped}
\begin{figure}[!ht]       %
  \centering
  \vspace{-3ex}
  \includegraphics[width=\linewidth]{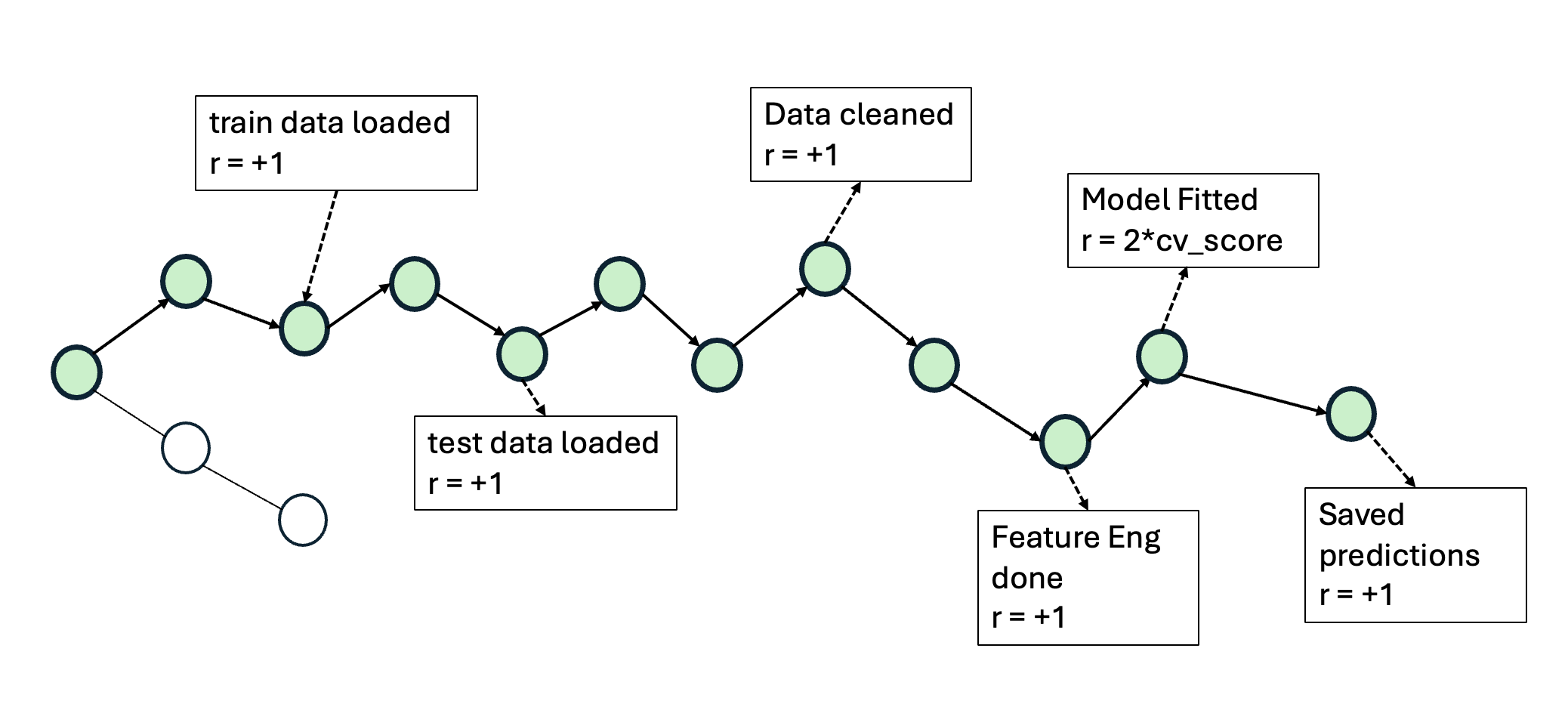} %
  \caption{An example illustration of how rewards are provided in MCTS-Shaped. If a particular stage is judged to be successfully completed at a node, a reward is given, which is used to update the value of all the nodes in this trajectory. It needs to be noted that these stage-wise rewards are only provided once per trajectory and only if the earlier stages were successfully completed.}
  \label{fig:rewards}
  \vspace{-3ex}
\end{figure}

In MCTS with shaped rewards, the agent receives intermediate credit for completing stages of the Kaggle ML challenge. The shaped-reward stages and their triggers are detailed below. Figure $\ref{fig:rewards}$ provides an example to illustrate how rewards are provided in MCTS-Shaped.

\noindent \textbf{Shaped-reward stages:}
\begin{enumerate}[leftmargin=*, nosep]
    \item \textbf{Train data loading}: reward when the agent successfully loads the training data.
    \item \textbf{Test data loading}: reward when the agent successfully loads the test data. Note that test data does not have the target variable, that needs to be predicted.
    \item \textbf{Combine train and test}: reward when the agent correctly concatenates train and test to enable consistent cleaning and feature engineering.
    \item \textbf{Data cleaning}: reward when no missing values (NaNs) remain in the combined data.
    \item \textbf{Feature engineering}: reward when (a) all categorical variables are properly encoded (e.g., one-hot or label encoding), and (b) the resulting feature dimensionality remains within a reasonable bound (to avoid exploding features from, e.g., high-cardinality text-like columns).
    \item \textbf{Split back to train/test}: reward when the agent correctly splits the combined data back into train and test after transformations.
    \item \textbf{Train features/target}: reward when the agent extracts $(X_{\text{train}}, y_{\text{train}})$ from the training dataframe using the correct target column.
    \item \textbf{Test features}: reward when the agent extracts $X_{\text{test}}$ from the test dataframe (which prior to this stage contains a dummy target), with correct arguments.
    \item \textbf{Modeling}: reward when the agent successfully fits a model on the training data; the reward is proportional to cross-validation performance.
    \item \textbf{Create submission}: reward when the agent generates predictions on the test data and writes a valid submission CSV to disk.
\end{enumerate}

It needs to be noted that all of the stage rewards are provided to the agent only once per trajectory, and only if the earlier stages were successfully completed. The provided stage rewards are used to update the value of all the nodes in the trajectory. We verify stage completion using a reward function that inspects the node scratchpad and tool messages, confirming (i) that artifacts satisfy required properties (e.g., no NaNs for data cleaning; all columns encoded for feature engineering) and (ii) that the correct tools were invoked as evidenced by the tool logs.

\subsection{Hierarchical MCTS}
\begin{figure}[H]       %
  \centering
  \vspace{-2ex}
  \includegraphics[width=\linewidth]{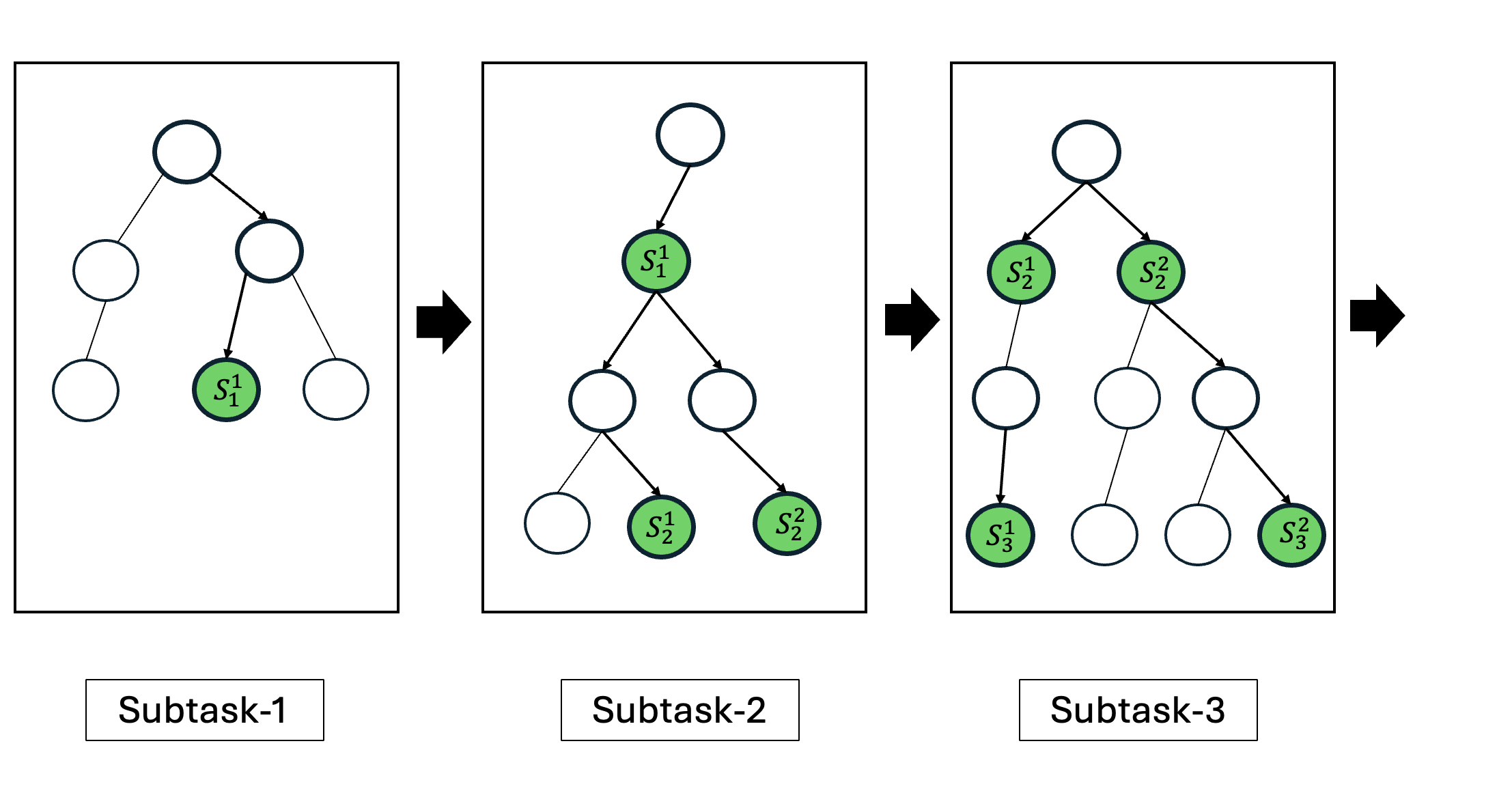} %
  \caption{A schematic of Hierarchical MCTS. The task is decomposed into an ordered sequence of subtasks. For each subtask, MCTS searches for all solution nodes up to a prescribed maximum subtask depth to avoid locally optimal but globally suboptimal choices. The solution nodes from subtask $t$ are appended to the root of subtask $t{+}1$, and the search resumes. In the example, the solution node from subtask~1, $S^1_{1}$, initializes subtask~2; its solution nodes $S^1_{2}$ and $S^2_{2}$ initialize subtask~3, and so on. The highest-value solution at the final subtask is returned as the overall outcome of Hierarchical MCTS.}
  \label{fig:hmcts}
  \vspace{-2ex}
\end{figure}
We propose Hierarchical MCTS to improve over ReAct \citep{yao2023reactsynergizingreasoningacting}, LATS \citep{zhou2024languageagenttreesearch}, and classical MCTS \citep{Kocsis2006BanditBM} in generating performant tool-use trajectories for solving Kaggle challenges within ML-Tool-Bench. Hierarchical MCTS decomposes a complex task into an ordered sequence of subtasks. In Hierarchical MCTS, we begin by decomposing the Kaggle challenge into a sequence of subtasks. This decomposition leverages the domain knowledge that solving a machine learning challenge typically involves data loading, data cleaning, feature engineering, modeling, evaluation, and prediction. We use a similar subtask decomposition to the reward-shaping structure used for MCTS-Shaped, described in Section~\ref{subsec:mcts-shaped}. Once we obtained textual descriptions for each subtask, a state-of-the-art coding agent was prompted with the subtask descriptions and the docstrings of the tools in our toolset, and asked to assign the relevant tools required to solve each subtask. One of the authors then manually reviewed the assignments to verify that the tool selections were sufficient and corrected minor errors made by the agent. This approach provides a general recipe for assigning tools to subtasks and can be applied in other domains, not only in the machine learning challenge–solving setting considered in this paper. 

For each subtask, MCTS searches its local state–action space to identify solution nodes. The solution nodes from one subtask are appended to the root of the next subtask, and the search continues. To avoid being trapped in locally optimal (but globally suboptimal) choices, we enumerate all solution nodes within each subtask up to a prescribed maximum subtask search depth. If there are no solution nodes identified after a subtask, the search terminates and we return `No Solution Found'. The solution node with the highest value, at the final subtask, is returned as the solution of the Hierarchical MCTS search. Note that, when solving for each subtask in Hierarchical MCTS, we do not use any reward shaping and only check for if the subtask was solved successfully or not. Importantly, the agent is given only the tools relevant to the current subtask (tool masking), which reduces the branching factor and focuses the search. Figure \ref{fig:hmcts} illustrates the overall procedure. Hierarchical MCTS is similar to the options framework \citep{sutton1999between}, that break down a complex problem into a hierarchy of sub-tasks, making the learning process more efficient and manageable.

\section{Experiments}
\label{sec: experiments}
We evaluate the tool-planning performance of two language models—GPT-4o and GPT-4.1-mini, on ML-Tool-Bench. For each model, we compare five planning algorithms: (i) \emph{ReAct} \citep{yao2023reactsynergizingreasoningacting}; (ii) \emph{LATS} \citep{zhou2024languageagenttreesearch}; (iii) Monte Carlo Tree Search (MCTS) with outcome-based rewards, where the agent is rewarded upon successfully training a model or producing a valid submission file (denoted \emph{MCTS-Outcome}); (iv) MCTS with shaped rewards, where the agent receives intermediate credit for completing stages of the Kaggle ML workflow (denoted \emph{MCTS-Shaped}); and (v) \emph{Hierarchical MCTS}: the Kaggle challenge is decomposed into subtasks. A node is a solution node for a subtask, if it satisfies the reward condition for the stage corresponding to that subtask.

\subsection{Implementation Details}
\begin{figure*}[t]
\vspace{-2ex}
    \centering
    \begin{subfigure}{0.48\textwidth}
        \centering
        \includegraphics[width=\textwidth]{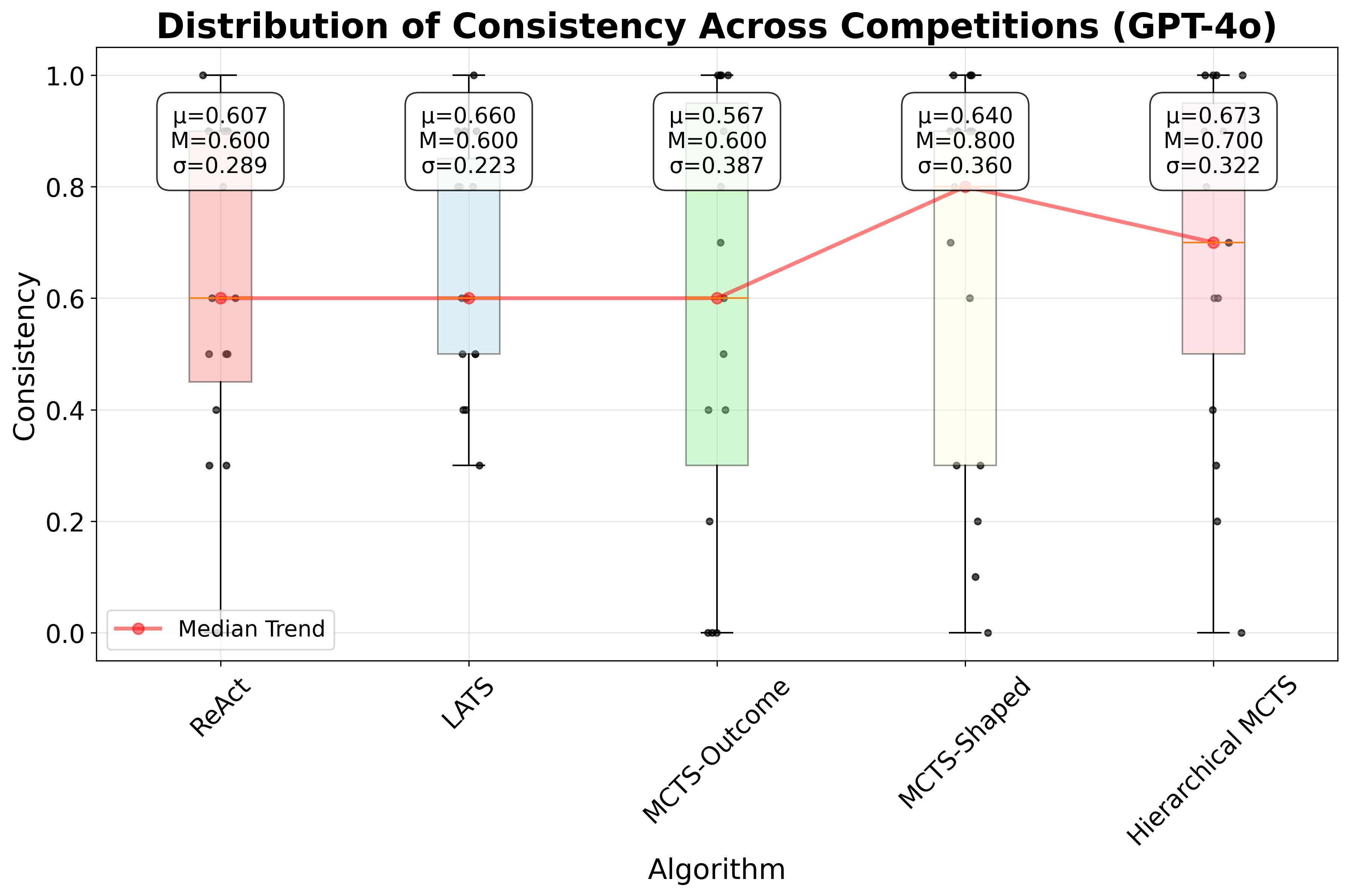}
        \label{fig:plot1}
    \end{subfigure}
    \hfill
    \begin{subfigure}{0.48\textwidth}
        \centering
        \includegraphics[width=\textwidth]{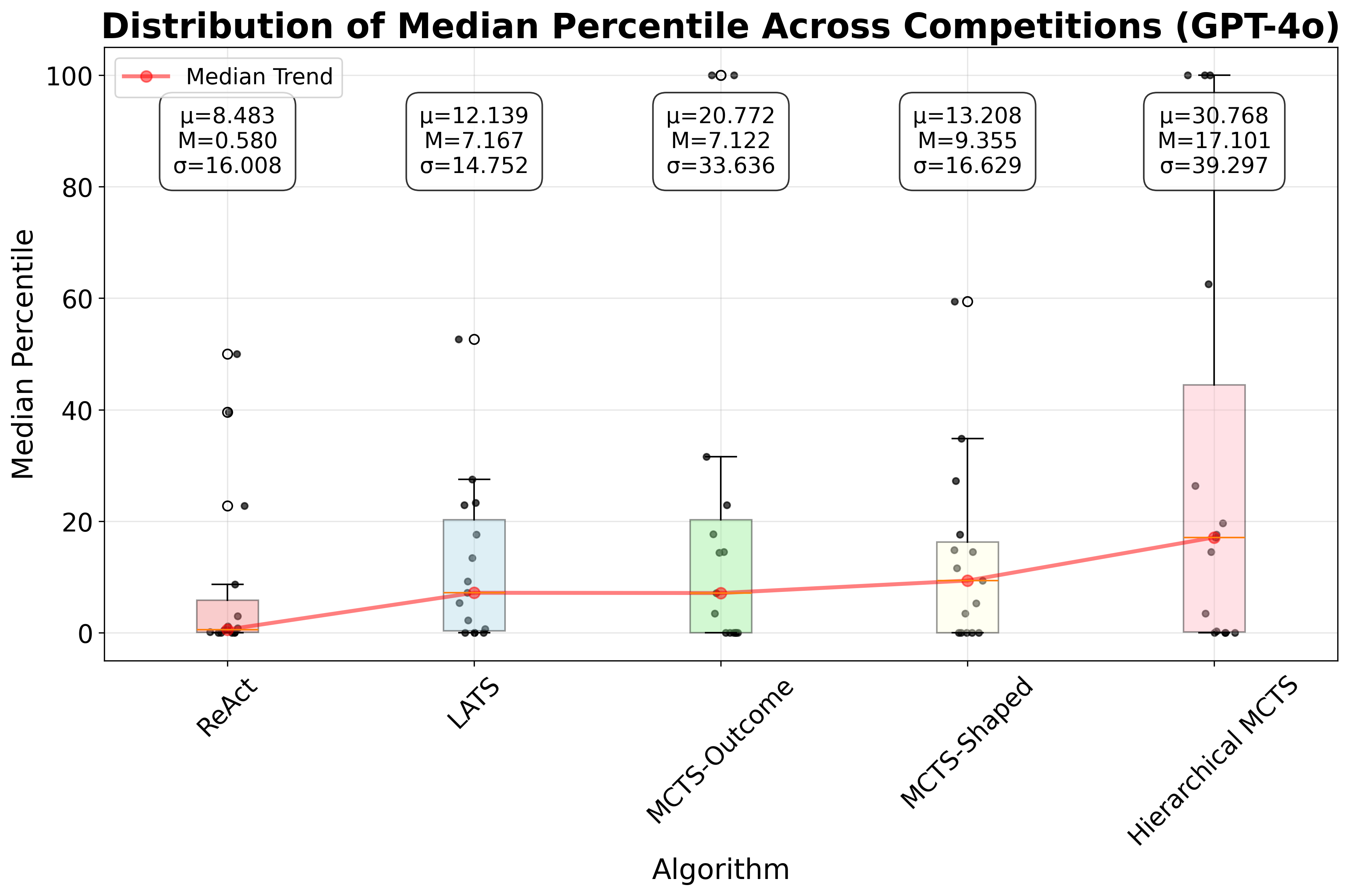}
        \label{fig:plot2}
    \end{subfigure}

    \begin{subfigure}{0.48\textwidth}
        \centering
        \includegraphics[width=\textwidth]{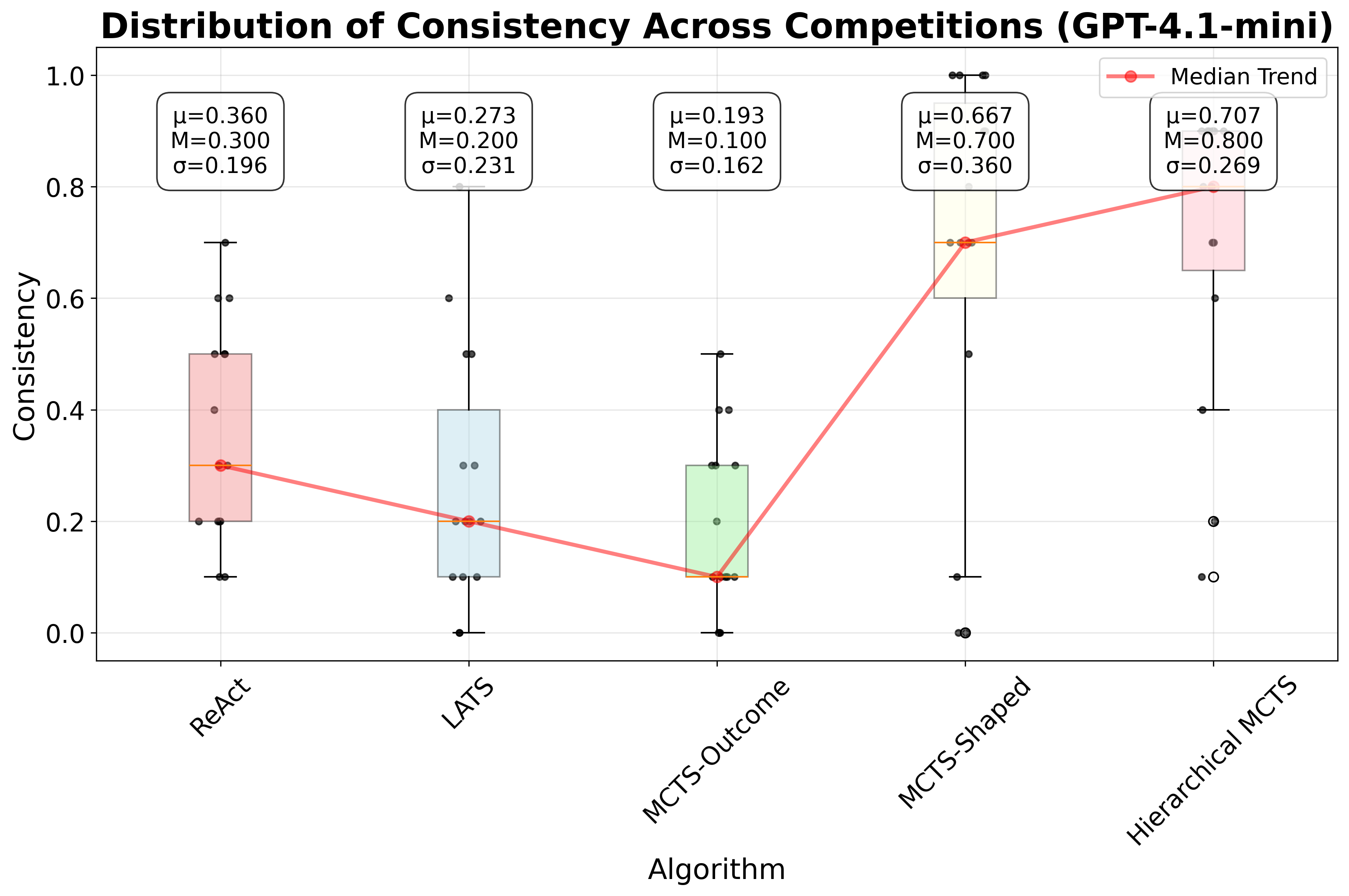}
        \label{fig:plot3}
    \end{subfigure}
    \hfill
    \begin{subfigure}{0.48\textwidth}
        \centering
        \includegraphics[width=\textwidth]{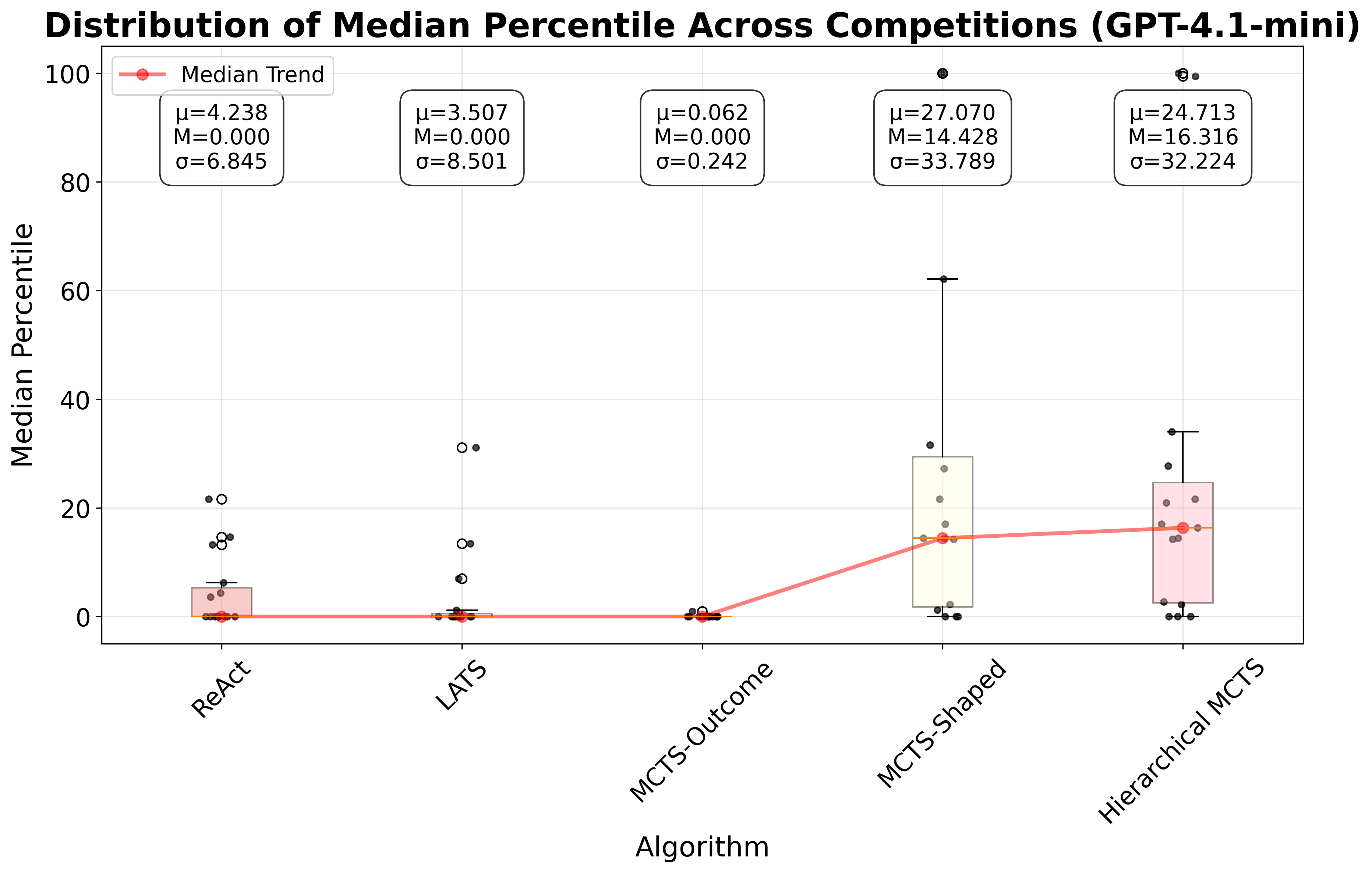}
        \label{fig:plot4}
    \end{subfigure}

    \caption{Plots of consistency and median leaderboard percentile across all competitions in ML-Tool-Bench, for different planning algorithms. The top row shows results for GPT-4o, and the bottom row shows results for GPT-4.1-mini. Hierarchical MCTS outperforms LATS and ReAct, followed by MCTS-Shaped, in terms of leaderboard performance for both LLMs. Both Hierarchical MCTS and MCTS-Shaped also improve consistency. In the box plots, $\mu$ denotes the mean, $\sigma$ the standard deviation, and $M$ the median.}
    \label{fig:boxplots}
    \vspace{-2ex}
\end{figure*}

When using tree-search methods with our in-memory, named-object scheme, we adopt a \emph{path-local scratchpad}, where each node $v$ contains a scratchpad $\mathcal{S}(v)$, that stores only the objects produced by the tool call at that node. During expansion, the LLM proposes candidate actions. For a candidate that is a tool call, the accessible memory is the path union: $
\mathcal{S}^{\ast}(v) \;=\; \bigcup_{u \in \mathrm{path}(\mathrm{root}\rightarrow v)} \mathcal{S}(u),
$
and the LLM may reference any named object in $\mathcal{S}^{\ast}(v)$ as tool arguments. The tool's outputs are written to the child's scratchpad $\mathcal{S}(\text{child})$, preserving isolation per node while enabling reuse of intermediate artifacts along the trajectory.

\textbf{LATS}: To estimate the value of a state, we provide an evaluator LLM with all \texttt{AIMessage} and \texttt{ToolMessage} entries along the path from the root to the current node. The evaluator scores the trajectory by the progress made toward solving the Kaggle challenge. To propose candidate actions, we similarly pass the full trajectory history to the LLM, which returns new reasoning steps or tool calls. Unlike original LATS, we omit the self-consistency score from the value estimate, as the agents were observed to typically propose a few distinct candidates per expansion, rendering a self-consistency score component unnecessary.

\textbf{MCTS}: We propose new candidate nodes during the expansion phase using the same approach as LATS. To evaluate the value of a node, we check if it produces a model or a valid submission file in the outcome rewards case. In the shaped rewards case, a node is provided a reward if it successfully completes a stage, as detailed earlier. In the case of Hierarchical MCTS, we designate a node as a solution node of the subtask, if it successfully completes the stage corresponding to that subtask. Additionally, across all MCTS variants, we apply a per-level depth penalty of 0.1 to discourage unnecessarily long trajectories that fail to make progress toward the goal.

In addition to rewards, we provide targeted textual feedback to help the agent refine its plan, when a stage fails. In such a case the agent receives a textual explanation of the failure. For example, in \emph{feature engineering} we flag existence of categorical columns or an excessive increase in feature dimensionality; in \emph{data cleaning} we report the presence of missing values. If a tool invocation fails, we return an explicit message along with the tool's docstring to guide correct usage on the next attempt. We point out that this tool docstring  is provided for all the planning approaches considered in the paper, upon tool invocation failure. We find that the textual feedback helps in producing valid trajectories more consistently. This textual feedback is provided for all the MCTS variants. 

Ideally, we would run Monte Carlo rollouts to a fixed depth or termination and use the return to update node values. However, running to termination is impractical, and even shallow rollouts (depth 3-5) across many challenges, algorithms, and trials proved cost-prohibitive. Learning value functions to approximate the value of states \citep{silver2016mastering} is also not straightforward, due to complex artifacts that are a part of the state. Consequently, we use the immediate reward at the current state (a depth-0 rollout), yielding a best-first search with a UCT-style exploration bonus. When budget permits, using small depth rollouts is preferred.

\subsection{Results}

We evaluate the tool-planning performance of two language models: GPT-4o and GPT-4.1-mini, on ML-Tool-Bench. We selected these models to balance capability and cost; frontier models such as GPT-5.2 were prohibitively expensive given our multi-trial, tree-search evaluation requiring hundreds of LLM calls per trajectory (see Appendix~\ref{ap:cost}). For each algorithm–Kaggle challenge combination, we run 10 trials. We define \emph{consistency} as the proportion of valid trajectories (e.g., 4 valid trajectories out of 10 trials yields a consistency of 0.4). For each trial, we evaluate predictions against the provided test labels using the competition’s official metric and compare against the leaderboard to obtain a leaderboard percentile. For each algorithm and competition, we report the median percentile across the 10 trials. Figure \ref{fig:boxplots} presents boxplots for all algorithms, summarizing the distribution of leaderboard percentiles across all competitions in our benchmark. For further details on consistency and leaderboard percentiles for both models, refer to Appendix $\ref{ap:results}$. For details on the prompts used, see Appendix $\ref{ap:prompts}$.
\vspace*{-1ex}
\paragraph{Benchmark Performance}
As shown in Figure~\ref{fig:boxplots}, Hierarchical MCTS improves leaderboard performance compared to ReAct, LATS, and MCTS-Outcome, followed by MCTS-Shaped, for both GPT-4o and GPT-4.1-mini. Moreover, both Hierarchical MCTS and MCTS-Shaped achieve higher consistency than the other baselines. For GPT-4o, Hierarchical MCTS shows improvement over LATS by 9.93 percentile positions on the leaderboard and over ReAct by 16.52 percentile positions, taking the median across all competitions. For GPT-4.1-mini, Hierarchical MCTS improved over MCTS-Shaped by 1.89 percentile positions on the leaderboard, while both ReAct and LATS had a median leaderboard percentile position of 0 across all competitions. 

These results highlight that as toolsets become more complex and larger, it is important either to introduce hierarchy—decomposing the original task into subtasks with corresponding reward functions, or to employ shaped rewards that guide the search toward solutions. In contrast, unidirectional planning strategies like ReAct do not perform well. Similarly, tree-search methods such as LATS, that rely solely on LLM evaluation also fail, as LLMs provide inconsistent scores to nodes when problem complexity and trajectory lengths increase—due to the accumulation of messages and artifacts that must be considered during evaluation. We also provide example trajectory visualizations in Appendix~\ref{ap:visualize}. An example successful trajectory produced by Hierarchical MCTS, also shown in Appendix~\ref{ap:visualize}, used 80 tool calls, explored multiple modeling strategies, and achieved a public leaderboard percentile of 62.65, whereas the other planning approaches did not explore multiple modeling strategies.

\begin{table*}[!ht]
\centering
\scriptsize
\caption{Consistency scores for the five planning approaches across six Kaggle challenges, evaluated using the original train/test splits provided by Kaggle.}
\label{tab:consistency-gpt-4.1-mini-kaggle-original}
\begin{tabular}{p{5.0cm} p{1.2cm} p{1.2cm} p{1.2cm} p{1.2cm} p{1.2cm}}
\toprule
\textbf{Competition} & \textbf{ReAct} & \textbf{LATS} & \textbf{MCTS-Outcome} & \textbf{MCTS-Shaped} & \textbf{Hierarchical MCTS} \\
\midrule
\href{https://www.kaggle.com/competitions/spaceship-titanic}{Spaceship Titanic} & 0.3 & 0.0 & \underline{\textbf{0.6}} & \underline{\textbf{0.6}} & 0.4 \\
\href{https://www.kaggle.com/competitions/playground-series-s5e9}{BPM Prediction} & 0.2 & 0.3 & 0 & \underline{\textbf{0.8}} & 0.7 \\
\href{https://www.kaggle.com/competitions/playground-series-s5e5}{Calorie Expenditure Prediction} & 0.4 & 0.3 & 0.2 & 0.3 & \underline{\textbf{0.6}} \\
\href{https://www.kaggle.com/competitions/playground-series-s3e1/overview}{California Housing Regression} & 0.6 & 0.4 & 0.1 & 0.5 & \underline{\textbf{0.8}} \\
\href{https://www.kaggle.com/competitions/playground-series-s5e8}{Bank Deposit Classification} & 0.4 & 0.5 & 0.0 & \underline{\textbf{1.0}} & 0.6 \\
\href{https://www.kaggle.com/competitions/playground-series-s4e1}{Bank Churn Classification} & 0.2 & 0.5 & 0.1 & 1.0 & \underline{\textbf{0.8}} \\
\midrule
\textbf{Overall (Median)} & 0.35 & 0.35 & 0.1 & \underline{\textbf{0.7}} & 0.65 \\
\bottomrule
\end{tabular}
\end{table*}

\begin{table*}[!ht]
\centering
\scriptsize
\caption{Median leaderboard percentiles for the five planning approaches across six Kaggle challenges. Percentiles are computed from Kaggle public leaderboard scores obtained via official submissions using the original train/test splits.}
\label{tab:leaderboard-percentile-gpt-4.1-mini-kaggle-original}
\begin{tabular}{p{5.0cm} p{1.2cm} p{1.2cm} p{1.2cm} p{1.2cm} p{1.2cm}}
\toprule
\textbf{Competition} & \textbf{ReAct} & \textbf{LATS} & \textbf{MCTS-Outcome} & \textbf{MCTS-Shaped} & \textbf{Hierarchical MCTS} \\
\midrule
\href{https://www.kaggle.com/competitions/spaceship-titanic}{Spaceship Titanic} & 0.0 & 0.0 & 39.77 & \underline{\textbf{41.69}} & 0.0 \\
\href{https://www.kaggle.com/competitions/playground-series-s5e9}{BPM Prediction} &  0 & 0 & 0 & \underline{\textbf{5.03}} & 0.19\\
\href{https://www.kaggle.com/competitions/playground-series-s5e5}{Calorie Expenditure Prediction} & 0.0 & 0.0 & 0.0 & 0.0 & \underline{\textbf{16.81}} \\
\href{https://www.kaggle.com/competitions/playground-series-s3e1/overview}{California Housing Regression} & 5.51 & 0.0 & 0.0 & 8.99 & \underline{\textbf{24.78}} \\
\href{https://www.kaggle.com/competitions/playground-series-s5e8}{Bank Deposit Classification} & 0.0 & 13.47 & 0.0 & \underline{\textbf{27.65}} & \underline{\textbf{27.29}} \\
\href{https://www.kaggle.com/competitions/playground-series-s4e1}{Bank Churn Classification} & 0.0 & 14.91 & 0.0 & 29.87 & \underline{\textbf{32.47}}\\
\midrule
\textbf{Overall (Median)} &  0.0 & 0.0 & 0.0 & 18.32 & \underline{\textbf{20.80}}\\
\bottomrule
\end{tabular}
\vspace{-4ex}
\end{table*}

\vspace*{-1ex}
\paragraph{Cost Analysis}
We analyze the computational costs of all planning algorithms considered in the paper and find that LATS is 3.5x more expensive than our stronger variants (Hierarchical MCTS and MCTS-Shaped), while achieving only 0.2 consistency and a median leaderboard percentile of 0 across five Kaggle challenges. In contrast, Hierarchical MCTS’s performance gains justify its computational cost: it outperforms both expensive but ineffective methods (LATS) and cheaper but poorly performing ones (ReAct), which costs 10.5x less than LATS yet achieves similarly near-zero performance. Additional cost details are provided in Appendix~\ref{ap:cost}.

\vspace*{-1ex}
\paragraph{Tool-Masking Ablation}
We further conduct a tool-masking ablation study (Appendix~\ref{ap:ablations}), which highlights the importance of tool masking in Hierarchical MCTS. Without tool masking, Hierarchical MCTS attains a median consistency of 0.3 and a median leaderboard percentile of 0 across five challenges; with tool masking, these improve to 0.8 and 21.10, respectively. This result indicates that both tool masking and subtask decomposition are critical for solving long-horizon planning problems in high-dimensional action spaces with LLMs.

\vspace*{-1ex}
\paragraph{Comparison to Direct Code-Generation}
We also compare our methods against a direct code-generation baseline, ReAct-Code, which produces end-to-end solutions and iteratively refines code using execution feedback. ReAct-Code outperforms all planning-based methods on both consistency and leaderboard percentile. Nevertheless, despite operating with a restricted toolset, our planning approaches remain competitive, suggesting that the chosen tools are largely sufficient for solving tabular Kaggle ML tasks. For more details, refer to Appendix ~\ref{ap:compare-code-gen}

\vspace*{-1ex}
\paragraph{ Kaggle Public Leaderboard Validation}
We then investigate if the benchmark performance serves as a reasonable proxy for the true Kaggle public leaderboard performance. For this analysis, we used the original Kaggle train and test splits rather than the smaller benchmark subsets. Since test labels were not available, we submitted our predictions to Kaggle to obtain the public leaderboard scores, which were then converted into percentile ranks. We note that our benchmark uses a subsampled portion of the training data and a test set partitioned from the training data, so labels are available. All other aspects remain unchanged. We report the Consistency and Leaderboard percentiles (with respect to the Kaggle public leaderboard) for the five planning approaches evaluated in this paper. Due to cost constraints, we evaluated only a subset of six challenges from our benchmark and used GPT-4.1-mini as the underlying LLM. The results are presented in the Tables ~\ref{tab:consistency-gpt-4.1-mini-kaggle-original} and ~\ref{tab:leaderboard-percentile-gpt-4.1-mini-kaggle-original}. We observe that these results exhibit the same trends as those seen in our benchmark evaluation: Hierarchical MCTS and MCTS-Shaped consistently outperform the other methods, while ReAct and LATS struggle, achieving a median leaderboard percentile of 0.0 across most of the six challenges evaluated. Hence, the benchmark performance serves as a reasonable proxy for the true Kaggle public leaderboard performance.

\section{Conclusion}
\label{sec: conclusion}

We introduced ML-Tool-Bench, the first benchmark for evaluating long-horizon planning in LLM agents on tabular Kaggle challenges, addressing a gap in existing benchmarks \citep{xu2023toolmanipulationcapabilityopensource, patil2025bfcl, yao2024taubenchbenchmarktoolagentuserinteraction} that focus on tool selection rather than multi-step planning. We found ReAct and LATS struggle to produce valid, performant trajectories, and proposed two improved approaches: MCTS with shaped deterministic rewards, and Hierarchical MCTS, which decomposes problems into sequenced subtasks. Across two models, Hierarchical MCTS achieved the best leaderboard performance while both methods improved trajectory validity, suggesting that subtask decomposition with deterministic rewards outperforms subjective LLM evaluation as tool complexity grows.

\section*{Impact Statement}
This paper presents work whose goal is to advance the field of Machine
Learning. There are many potential societal consequences of our work, none
which we feel must be specifically highlighted here.

\bibliography{example_paper}
\bibliographystyle{icml2026}

\newpage
\appendix
\onecolumn
\label{sec: appendix}
\section{Approaches}\label{ap:approaches}

\subsection{Monte Carlo Tree Search}

\begin{figure}[h]       %
  \centering
  \includegraphics[width=0.8\linewidth]{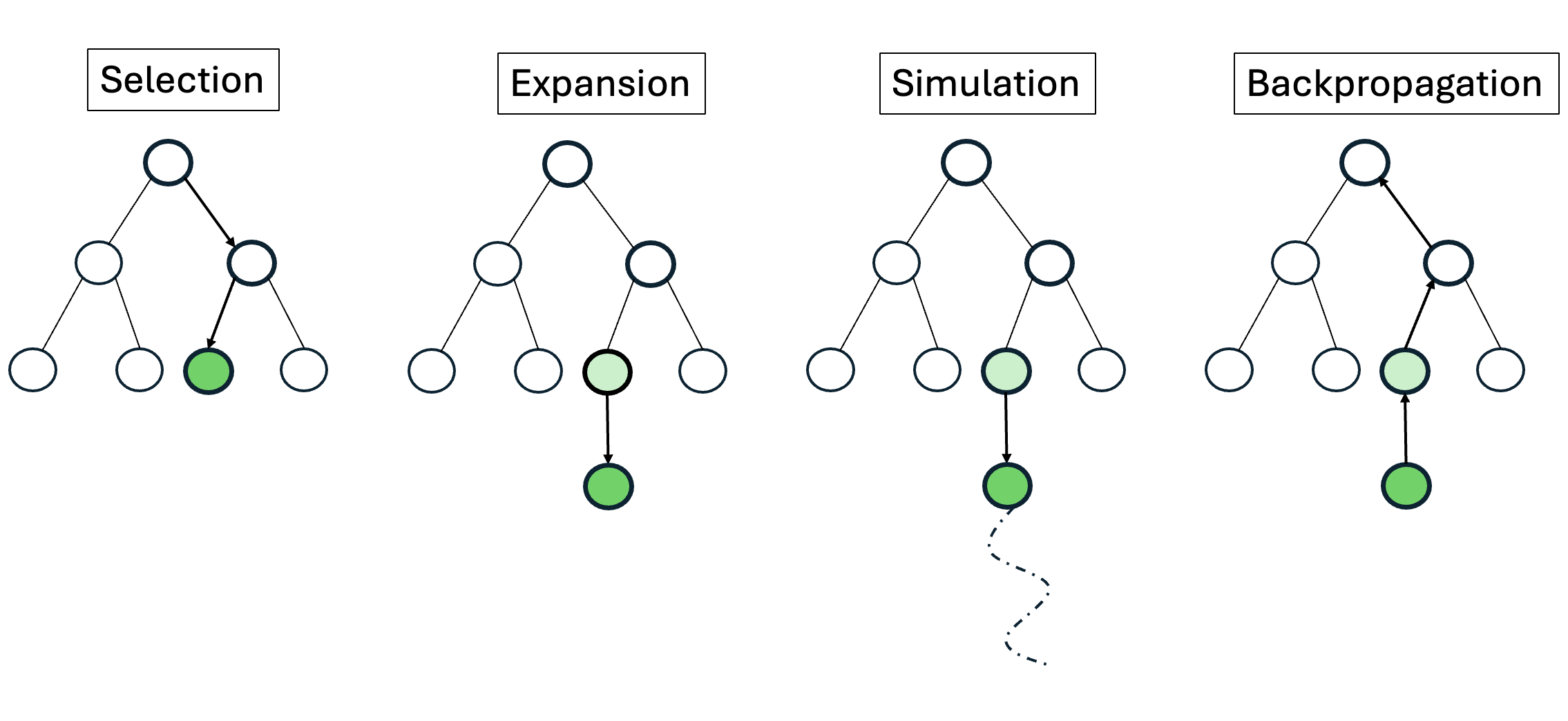} %
  \caption{A pictorial illustration of Monte Carlo Tree Search}
  \label{fig:mcts}
\end{figure}

Figure $\ref{fig:mcts}$ provides a pictorial illustration of the MCTS algorithm.

\section{Kaggle Challenges}\label{ap:kaggle}

The list of Kaggle challenges present in ML-Tool-Bench, and the corresponding ML problem types of each challenge are presented in Table $\ref{tab:challenges}$

\begin{table}[ht]
\centering
\caption{Kaggle challenges used in ML-Tool-Bench with problem type.}
\label{tab:challenges}
\begin{tabular}{ll}
\hline
\textbf{Challenge} & \textbf{Type} \\
\hline
\href{https://www.kaggle.com/competitions/santander-value-prediction-challenge/overview}{Santander Value Prediction Challenge} & Regression \\
\href{https://www.kaggle.com/competitions/new-york-city-taxi-fare-prediction}{New York City Taxi Fare Prediction} & Regression \\
\href{https://www.kaggle.com/competitions/nyc-taxi-trip-duration}{New York City Taxi Trip Duration} & Regression \\
\href{https://www.kaggle.com/competitions/playground-series-s5e9}{Predicting the Beats-per-Minute of Songs} & Regression \\
\href{https://www.kaggle.com/competitions/playground-series-s5e5}{Predict Calorie Expenditure} & Regression \\
\href{https://www.kaggle.com/competitions/playground-series-s3e1/overview}{Regression with a Tabular California Housing Dataset} & Regression \\
\href{https://www.kaggle.com/competitions/playground-series-s4e9}{Regression of Used Car Prices} & Regression \\
\hline
\href{https://www.kaggle.com/competitions/porto-seguro-safe-driver-prediction}{Porto Seguro Safe Driver Prediction} & Binary Classification \\
\href{https://www.kaggle.com/c/costa-rican-household-poverty-prediction}{Costa Rican Household Poverty Prediction} & Multi-Class Classification \\
\href{https://www.kaggle.com/competitions/forest-cover-type-kernels-only}{Forest Cover Type (Kernels Only)} & Multi-Class Classification \\
\href{https://www.kaggle.com/competitions/santander-customer-transaction-prediction}{Santander Customer Transaction Prediction} & Binary Classification \\
\href{https://www.kaggle.com/competitions/playground-series-s4e8}{Binary Prediction of Poisonous Mushrooms} & Binary Classification \\
\href{https://www.kaggle.com/competitions/spaceship-titanic}{Spaceship Titanic} & Binary Classification \\
\href{https://www.kaggle.com/competitions/playground-series-s5e8}{Binary Classification with a Bank Dataset} & Binary Classification \\
\href{https://www.kaggle.com/competitions/playground-series-s4e1}{Binary Classification with a Bank Churn Dataset} & Binary Classification \\
\hline
\end{tabular}
\end{table}

\section{Results}\label{ap:results}
In this section, we provide the exact consistency and performance values for each of the 15 challenges and the two models (GPT-4o and GPT-4.1-mini). Tables~\ref{tab:consistency-gpt-4o} and~\ref{tab:percentile-gpt-4o} show the consistency and leaderboard percentiles for all algorithms across all competitions in ML-Tool-Bench, for GPT-4o. Similarly, Tables~\ref{tab:consistency-gpt-4.1-mini} and~\ref{tab:percentile-gpt-4.1-mini} show the consistency and leaderboard percentiles for all algorithms across all competitions in ML-Tool-Bench, for GPT-4.1-mini.
\begin{table}[t]
\centering
\scriptsize  %
\caption{Consistency across 15 competitions for five planning algorithms for GPT-4o.}
\label{tab:consistency-gpt-4o}
\begin{tabular}{p{5.0cm} p{1.2cm} p{1.2cm} p{1.2cm} p{1.2cm} p{1.2cm}}
\toprule
\textbf{Competition} & \textbf{ReAct} & \textbf{LATS} & \textbf{MCTS-Outcome} & \textbf{MCTS-Shaped} & \textbf{Hierarchical MCTS} \\
\midrule
\href{https://www.kaggle.com/competitions/spaceship-titanic}{Spaceship Titanic} & 0.6 & 0.4 & \underline{\textbf{0.9}} & 0.8 & 0.6 \\
\href{https://www.kaggle.com/competitions/santander-value-prediction-challenge/overview}{Santander Value Prediction Challenge} & 0.9 & 0.6 & 0.5 & \underline{\textbf{1}} & \underline{\textbf{1}} \\
\href{https://www.kaggle.com/competitions/new-york-city-taxi-fare-prediction}{NYC Taxi Fare Prediction} & 0 & 0.5 & 0.2 & 0.3 & \underline{\textbf{0.9}} \\
\href{https://www.kaggle.com/competitions/nyc-taxi-trip-duration}{NYC Taxi Trip Duration} & 0.3 & 0.3 & \underline{\textbf{1}} & 0.2 & 0.3 \\
\href{https://www.kaggle.com/competitions/playground-series-s5e9}{BPM Prediction} & \underline{\textbf{1}} & 0.6 & 0 & 0.9 & 0.7 \\
\href{https://www.kaggle.com/competitions/playground-series-s5e5}{Calorie Expenditure Prediction} & 0.8 & 0.8 & \underline{\textbf{1}} & 0.9 & \underline{\textbf{1}} \\
\href{https://www.kaggle.com/competitions/playground-series-s3e1/overview}{california Housing Regression} & 0.9 & 0.9 & \underline{\textbf{1}} & 0.9 & 0.9 \\
\href{https://www.kaggle.com/competitions/playground-series-s4e9}{Used Car Prices Regression} & 0.9 & 0.4 & 0.4 & 0.9 & \underline{\textbf{1}} \\
\href{https://www.kaggle.com/competitions/porto-seguro-safe-driver-prediction}{Porto Seguro Safe Driver Prediction} & 0.3 & \underline{\textbf{0.5}} & 0 & 0.1 & 0.2 \\
\href{https://www.kaggle.com/c/costa-rican-household-poverty-prediction}{Costa Rican Household Poverty Level Prediction} & 0.5 & 0.5 & 0.7 & 0.3 & \underline{\textbf{1}} \\
\href{https://www.kaggle.com/competitions/forest-cover-type-kernels-only}{Forest Cover Type Prediction} & 0.6 & \underline{\textbf{0.9}} & 0 & 0 & 0 \\
\href{https://www.kaggle.com/competitions/santander-customer-transaction-prediction}{Santander Customer Transaction Prediction} & 0.5 & \underline{\textbf{0.8}} & \underline{\textbf{0.8}} & 0.7 & 0.4 \\
\href{https://www.kaggle.com/competitions/playground-series-s4e8}{Poisonous Mushroom Prediction} & 0.9 & \underline{\textbf{1}} & \underline{\textbf{1}} & \underline{\textbf{1}} & 0.8 \\
\href{https://www.kaggle.com/competitions/playground-series-s5e8}{Bank Deposit Classification} & 0.5 & \underline{\textbf{0.8}} & 0.4 & 0.6 & 0.7 \\
\href{https://www.kaggle.com/competitions/playground-series-s4e1}{Bank Churn Classification} & 0.4 & 0.9 & 0.6 & \underline{\textbf{1}} & 0.6 \\
\midrule
\textbf{Overall (Median)} & 0.6 & 0.6 & 0.6 & \underline{\textbf{0.8}} & 0.7 \\
\bottomrule
\end{tabular}
\end{table}

\begin{table}[t]
\centering
\scriptsize  %
\caption{Median Leaderboard percentile across 15 competitions for five planning algorithms for GPT-4o.}
\label{tab:percentile-gpt-4o}
\begin{tabular}{p{5.0cm} p{1.2cm} p{1.2cm} p{1.2cm} p{1.2cm} p{1.2cm}}
\toprule
\textbf{Competition} & \textbf{ReAct} & \textbf{LATS} & \textbf{MCTS-Outcome} & \textbf{MCTS-Shaped} & \textbf{Hierarchical MCTS} \\
\midrule
\href{https://www.kaggle.com/competitions/spaceship-titanic}{Spaceship Titanic} & 39.54 & 0 &	22.88	& 59.44 &	\underline{\textbf{62.55}} \\
\href{https://www.kaggle.com/competitions/santander-value-prediction-challenge/overview}{Santander Value Prediction Challenge} & 0.09 & 7.17 & 7.12 & \underline{\textbf{14.87}} & 0.27 \\
\href{https://www.kaggle.com/competitions/nyc-taxi-trip-duration}{NYC Taxi Fare Prediction} & 0 & 9.23 &	0 &	0 &	\underline{\textbf{19.66}} \\
\href{https://www.kaggle.com/competitions/nyc-taxi-trip-duration}{NYC Taxi Trip Duration} & 0 & 0 & \underline{\textbf{100.0}} & 0 & 0 \\
\href{https://www.kaggle.com/competitions/playground-series-s5e9}{BPM Prediction} & 0.51 & 52.63 & 0 & 5.26 & \underline{\textbf{100}}\\
\href{https://www.kaggle.com/competitions/playground-series-s5e5}{Calorie Expenditure Prediction} & 0.16 & 13.43 & \underline{\textbf{14.47}} & \underline{\textbf{14.47}} & \underline{\textbf{14.47}} \\
\href{https://www.kaggle.com/competitions/playground-series-s3e1/overview}{california Housing Regression} & 0.58 & 0.65 & 14.35 & 11.59 &	\underline{\textbf{17.10}} \\
\href{https://www.kaggle.com/competitions/playground-series-s4e9}{Used Car Prices Regression} & 3.0 &	0 & 0 & 9.35 & \underline{\textbf{100}} \\
\href{https://www.kaggle.com/competitions/porto-seguro-safe-driver-prediction}{Porto Seguro Safe Driver Prediction} & 0 & \underline{\textbf{5.34}} & 0 & 0 & 0 \\
\href{https://www.kaggle.com/c/costa-rican-household-poverty-prediction}{Costa Rican Household Poverty Level Prediction} & 50 & 0 & \underline{\textbf{100}} & 0 & \underline{\textbf{100}} \\
\href{https://www.kaggle.com/competitions/forest-cover-type-kernels-only}{Forest Cover Type Prediction} & 0.84 & \underline{\textbf{23.32}} & 0 & 0 & 0\\
\href{https://www.kaggle.com/competitions/santander-customer-transaction-prediction}{Santander Customer Transaction Prediction} & 1.14 & 2.27	& \underline{\textbf{3.49}} & \underline{\textbf{3.49}} & 0 \\
\href{https://www.kaggle.com/competitions/playground-series-s4e8}{Poisonous Mushroom Prediction} & \underline{\textbf{22.75}} & 17.61 & 17.69 & 17.62 & 17.62\\
\href{https://www.kaggle.com/competitions/playground-series-s5e8}{Bank Deposit Classification} & 8.64	& \underline{\textbf{27.50}} & 0 & 27.24 & 26.37 \\
\href{https://www.kaggle.com/competitions/playground-series-s4e1}{Bank Churn Classification} & 0	& 22.93 & 31.57 & \underline{\textbf{34.79}} & 3.47 \\
\midrule
\textbf{Overall (Median)} & 0.58 & 	7.17 & 7.12  & 9.36 & \underline{\textbf{17.10}} \\
\bottomrule
\end{tabular}
\end{table}

\begin{table}[!h]
\centering
\scriptsize  %
\caption{Consistency across 15 competitions for five planning algorithms for GPT-4.1-mini.}
\label{tab:consistency-gpt-4.1-mini}
\begin{tabular}{p{5.0cm} p{1.2cm} p{1.2cm} p{1.2cm} p{1.2cm} p{1.2cm}}
\toprule
\textbf{Competition} & \textbf{ReAct} & \textbf{LATS} & \textbf{MCTS-Outcome} & \textbf{MCTS-Shaped} & \textbf{Hierarchical MCTS} \\
\midrule
\href{https://www.kaggle.com/competitions/spaceship-titanic}{Spaceship Titanic} & 0.1 & 0 & 0.4 & \underline{\textbf{0.7}} & 0.2 \\
\href{https://www.kaggle.com/competitions/santander-value-prediction-challenge/overview}{Santander Value Prediction Challenge} & 0.6 & 0.2 & 0 & \underline{\textbf{0.9}} & \underline{\textbf{0.9}} \\
\href{https://www.kaggle.com/competitions/new-york-city-taxi-fare-prediction}{NYC Taxi Fare Prediction} & 0.2 & 0 & 0.1 & 0.1 & \underline{\textbf{0.9}} \\
\href{https://www.kaggle.com/competitions/nyc-taxi-trip-duration}{NYC Taxi Trip Duration} & 0.1 & 0.1 & 0.3 & 0.5 & \underline{\textbf{0.9}} \\
\href{https://www.kaggle.com/competitions/playground-series-s5e9}{BPM Prediction} & 0.4 & 0.1 & 0 & \underline{\textbf{1}} & 0.8 \\
\href{https://www.kaggle.com/competitions/playground-series-s5e5}{Calorie Expenditure Prediction} & 0.6 & 0.6 & 0 & \underline{\textbf{0.9}} & \underline{\textbf{0.9}} \\
\href{https://www.kaggle.com/competitions/playground-series-s3e1/overview}{california Housing Regression} & 0.7 & 0.3 & 0.3 & \underline{\textbf{0.8}} & \underline{\textbf{0.8}} \\
\href{https://www.kaggle.com/competitions/playground-series-s4e9}{Used Car Prices Regression} & 0.3 & 0.2 &	0.4 & \underline{\textbf{0.7}} & 0.4 \\
\href{https://www.kaggle.com/competitions/porto-seguro-safe-driver-prediction}{Porto Seguro Safe Driver Prediction} & 0.2 & 0.1 & 0.1 & 0.7 & \underline{\textbf{0.9}} \\
\href{https://www.kaggle.com/c/costa-rican-household-poverty-prediction}{Costa Rican Household Poverty Level Prediction} & \underline{\textbf{0.2}} & \underline{\textbf{0.2}} & 0.1 & 0 & 0.1 \\
\href{https://www.kaggle.com/competitions/forest-cover-type-kernels-only}{Forest Cover Type Prediction} & 0.2 & 0.2 & 0.3 & 0 & \underline{\textbf{0.7}} \\
\href{https://www.kaggle.com/competitions/santander-customer-transaction-prediction}{Santander Customer Transaction Prediction} & 0.3 & 0.5 & 0.1 & \underline{\textbf{0.7}} & \underline{\textbf{0.7}} \\
\href{https://www.kaggle.com/competitions/playground-series-s4e8}{Poisonous Mushroom Prediction} & 0.5 & 0.5 &	0.2	& \underline{\textbf{1}} &	0.6 \\
\href{https://www.kaggle.com/competitions/playground-series-s5e8}{Bank Deposit Classification} & 0.5 & 0.3 & 0.1 & 	\underline{\textbf{1}} &	0.9 \\
\href{https://www.kaggle.com/competitions/playground-series-s4e1}{Bank Churn Classification} & 0.5 & 0.8 & 0.5 & \underline{\textbf{1}} &	0.9 \\
\midrule
\textbf{Overall (Median)} & 0.3 &	0.2&	0.1	& 0.7 & \underline{\textbf{0.8}} \\
\bottomrule
\end{tabular}
\end{table}

\begin{table}[t]
\centering
\scriptsize  %
\caption{Median Leaderboard percentile across 15 competitions for five planning algorithms for GPT-4.1-mini.}
\label{tab:percentile-gpt-4.1-mini}
\begin{tabular}{p{5.0cm} p{1.2cm} p{1.2cm} p{1.2cm} p{1.2cm} p{1.2cm}}
\toprule
\textbf{Competition} & \textbf{ReAct} & \textbf{LATS} & \textbf{MCTS-Outcome} & \textbf{MCTS-Shaped} & \textbf{Hierarchical MCTS} \\
\midrule
\href{https://www.kaggle.com/competitions/spaceship-titanic}{Spaceship Titanic} & 0	& 0	& 0	& \underline{\textbf{62.11}}	& 0 \\
\href{https://www.kaggle.com/competitions/santander-value-prediction-challenge/overview}{Santander Value Prediction Challenge} & 3.60	& 0 & 	0 &	\underline{\textbf{14.24}} &	\underline{\textbf{14.24}} \\
\href{https://www.kaggle.com/competitions/new-york-city-taxi-fare-prediction}{NYC Taxi Fare Prediction} & 0	& 0	& 0	& 0	& \underline{\textbf{20.94}} \\
\href{https://www.kaggle.com/competitions/nyc-taxi-trip-duration}{NYC Taxi Trip Duration} & 0 & 0 & 0 &	1.24 & 	\underline{\textbf{2.70}} \\
\href{https://www.kaggle.com/competitions/playground-series-s5e9}{BPM Prediction} & 0 & 0 & 0 & \underline{\textbf{100}} & \underline{\textbf{100}}\\
\href{https://www.kaggle.com/competitions/playground-series-s5e5}{Calorie Expenditure Prediction} & 4.31	& 13.41 & 0 & \underline{\textbf{14.43}} & \underline{\textbf{14.43}}  \\
\href{https://www.kaggle.com/competitions/playground-series-s3e1/overview}{california Housing Regression} & \underline{\textbf{21.59}} &	0 &	0 &	\underline{\textbf{21.59}} &	\underline{\textbf{21.59}} \\
\href{https://www.kaggle.com/competitions/playground-series-s4e9}{Used Car Prices Regression} & 0 & 0 & 0	& \underline{\textbf{100}} &	0 \\
\href{https://www.kaggle.com/competitions/porto-seguro-safe-driver-prediction}{Porto Seguro Safe Driver Prediction} & 0	& 0	& 0	& \underline{\textbf{17.01}} &	\underline{\textbf{17.01}} \\
\href{https://www.kaggle.com/c/costa-rican-household-poverty-prediction}{Costa Rican Household Poverty Level Prediction} & 0	& 0 & 0 & 0 & 0 \\
\href{https://www.kaggle.com/competitions/forest-cover-type-kernels-only}{Forest Cover Type Prediction} & 0 &	0 & 0 & 0 & \underline{\textbf{99.44}}\\
\href{https://www.kaggle.com/competitions/santander-customer-transaction-prediction}{Santander Customer Transaction Prediction} & 0	& 1.17	& 0 & \underline{\textbf{2.27}} &	\underline{\textbf{2.27}} \\
\href{https://www.kaggle.com/competitions/playground-series-s4e8}{Poisonous Mushroom Prediction} & 6.23 & 	6.97 & 0 & 14.36 &	\underline{\textbf{16.32}}\\
\href{https://www.kaggle.com/competitions/playground-series-s5e8}{Bank Deposit Classification} & 13.19 &	0 &	0	& 27.24 &	\underline{\textbf{27.73}} \\
\href{https://www.kaggle.com/competitions/playground-series-s4e1}{Bank Churn Classification} & 14.66 &	31.09 &	0.94 &	31.57 &	\underline{\textbf{34.02}} \\
\midrule
\textbf{Overall (Median)} & 0 & 0 & 0 & 14.43 &	\underline{\textbf{16.32}} \\
\bottomrule
\end{tabular}
\end{table}

\pagebreak

\section{Tools}\label{ap:tools}
In this section, we describe the various tools that are part of ML-Tool-Bench. Table~\ref{tab:tools-workflow} shows the number of tools in our toolset that are part of each stage in solving an ML challenge on Kaggle. Table~\ref{tab:all-tools} provides info about all the tools in the curated toolset provided by ML-Tool-Bench
\begin{table}[t]
\centering
\caption{Number of tools available at each stage of a Kaggle-style workflow. In total, 61 tools are provided spanning data loading, cleaning, feature engineering, and modeling. Some tools can appear in more than one stage}
\label{tab:tools-workflow}
\begin{tabular}{l c}
\toprule
\textbf{Stage}              & \textbf{Number of Tools} \\
\midrule
Data Loading                & 6  \\
Data Cleaning               & 9 \\
Feature Engineering         & 30 \\
Modeling & 10 \\
Evaluation/Prediction & 10 \\
\bottomrule
\end{tabular}
\end{table}

\paragraph{Decorators for named references}
To enable tools to operate on named references rather than raw objects, we design four decorators that adapt arbitrary user-provided functions to our scratchpad interface according to their read--write behavior. We categorize tools into four types:
\begin{enumerate}
    \item \textbf{Set tool}: saves an object to memory. Example: \texttt{read\_csv} loads a dataframe and stores it under a provided name.
    \item \textbf{Get tool}: reads an object from memory. Example: \texttt{get\_dataframe\_summary} loads a dataframe and returns a brief textual summary to guide subsequent planning.
    \item \textbf{Get--Set tool}: reads an object from memory and writes a new object to memory. Example: \texttt{fit\_randomforest\_model} takes as input, a dataframe, and returns a fitted model.
    \item \textbf{Override tool}: reads an object, returns an updated object, and overwrites the input variable binding with the returned value. Example: \texttt{cast\_column} loads a dataframe and returns a modified dataframe that replaces the original.
\end{enumerate}
Accordingly, we provide four decorators: \texttt{make\_get\_tool}, \texttt{make\_set\_tool}, \texttt{make\_get\_and\_set\_tool}, and \texttt{make\_override\_tool}, that automatically wrap user-provided tools to operate on named references and integrate with the scratchpad.

\begin{longtable}{|p{8cm}|p{6cm}|}
\caption{All tools in the curated toolset provided by ML-Tool-Bench.}
\label{tab:all-tools}\\
\hline
\textbf{Function Signature} & \textbf{Description} \\
\hline

\multicolumn{2}{|c|}{\textbf{Modeling Functions}} \\
\hline
\texttt{fit\_logistic\_regressor(X\_train, y\_train, cv=5)} & Fit Logistic Regression model \\
\hline
\texttt{fit\_linear\_regressor(X\_train, y\_train, cv=5)} & Fit Linear Regression model \\
\hline
\texttt{fit\_random\_forest\_regressor(X\_train, y\_train, cv=5)} & Fit Random Forest Regressor \\
\hline
\texttt{fit\_random\_forest\_classifier(X\_train, y\_train, cv=5)} & Fit Random Forest Classifier \\
\hline
\texttt{fit\_xgboost\_regressor(X\_train, y\_train, cv=5)} & Fit XGBoost Regressor \\
\hline
\texttt{fit\_xgboost\_classifier(X\_train, y\_train, cv=5)} & Fit XGBoost Classifier  \\
\hline
\texttt{fit\_lightgbm\_regressor(X\_train, y\_train, cv=5)} & Fit LightGBM Regressor \\
\hline
\texttt{fit\_lightgbm\_classifier(X\_train, y\_train, cv=5)} & Fit LightGBM Classifier \\
\hline
\texttt{fit\_catboost\_regressor(X\_train, y\_train, cv=5)} & Fit CatBoost Regressor \\
\hline
\texttt{fit\_catboost\_classifier(X\_train, y\_train, cv=5)} & Fit CatBoost Classifier \\
\hline

\multicolumn{2}{|c|}{\textbf{Data Loading Functions}} \\
\hline
\texttt{read\_data(filepath)} & Read CSV data into a pandas DataFrame \\
\hline

\multicolumn{2}{|c|}{\textbf{Feature Engineering and Functions to get Dataframe information}} \\
\hline
\texttt{create\_numeric\_feature(df, name, expression)} & Create a numeric feature using a pandas expression \\
\hline
\texttt{create\_categorical\_feature(df, name, source\_column, mapping)} & Create a categorical feature by mapping values from a source column \\
\hline
\texttt{create\_conditional\_feature(df, name, condition, true\_value, false\_value)} & Create a feature based on a condition \\
\hline
\texttt{extract\_string\_pattern(df, name, source\_column, pattern, group=0)} & Extract pattern from string column using regex \\
\hline
\texttt{split\_string\_column(df, name\_prefix, source\_column, delimiter, max\_splits=-1, indices=None)} & Split string column and create separate features \\
\hline
\texttt{create\_group\_aggregation(df, name, group\_column, agg\_column, agg\_func)} & Create feature by aggregating within groups \\
\hline
\texttt{get\_group\_aggregation(df, group\_column, agg\_column, agg\_func)} & Get aggregation result without adding it to the DataFrame \\
\hline
\texttt{create\_rolling\_feature(df, name, source\_column, window, agg\_func='mean')} & Create rolling window feature \\
\hline
\texttt{create\_lag\_feature(df, name, source\_column, lag=1)} & Create lagged feature \\
\hline
\texttt{create\_lead\_feature(df, name, source\_column, lead=1)} & Create leading feature \\
\hline
\texttt{extract\_datetime\_features(df, datetime\_column, features=None)} & Extract datetime features from datetime column \\
\hline
\texttt{create\_time\_delta(df, name, start\_column, end\_column, unit='D')} & Create time delta feature between two datetime columns \\
\hline
\texttt{apply\_custom\_function(df, name, source\_columns, func)} & Apply custom function to create feature \\
\hline
\texttt{fillna\_with\_value(df, columns, value)} & Fill missing values with a specific value \\
\hline
\texttt{fillna\_with\_median(df, columns=None)} & Fill missing values with median of the column \\
\hline
\texttt{fillna\_with\_mean(df, columns=None)} & Fill missing values with mean of the column \\
\hline
\texttt{fillna\_with\_mode(df, columns=None)} & Fill missing values with mode of the column \\
\hline
\texttt{fillna\_with\_condition(df, target\_column, condition, fill\_value)} & Fill missing values in a column based on a condition \\
\hline
\texttt{fillna\_with\_multiple\_conditions(df, target\_column, conditions\_and\_values)} & Fill missing values in a column based on multiple conditions \\
\hline
\texttt{fillna\_with\_conditional\_aggregation(df, target\_column, condition\_column, condition\_values, agg\_func='mean')} & Fill missing values using conditional aggregation based on another column's values \\
\hline
\texttt{fillna\_with\_custom\_function(df, target\_column, condition, custom\_func)} & Fill missing values using a custom function based on a condition \\
\hline
\texttt{drop\_rows\_with\_missing(df, columns=None, threshold=None)} & Drop rows with missing values \\
\hline
\texttt{get\_missing\_summary(df)} & Get a summary of missing values in the DataFrame \\
\hline
\texttt{cast\_columns(df, column\_type\_mapping)} & Cast columns to specified data types \\
\hline
\texttt{cast\_numeric\_columns(df, columns=None, target\_type='float')} & Cast numeric columns to specified type \\
\hline
\texttt{cast\_integer\_columns\_to\_float(df, columns=None)} & Cast integer columns to float type \\
\hline
\texttt{cast\_categorical\_columns(df, columns=None)} & Cast categorical columns to category type \\
\hline
\texttt{one\_hot\_encode(df, columns=None, drop\_first=True, prefix=None)} & One-hot encode categorical columns \\
\hline
\texttt{label\_encode(df, columns=None)} & Label encode categorical columns \\
\hline
\texttt{normalize\_features(df, columns=None, method='standard')} & Normalize numeric features \\
\hline
\texttt{encode\_all\_categorical\_columns(df, method='one\_hot', drop\_first=True)} & Encode all categorical/object columns using specified method \\
\hline
\texttt{normalize\_all\_numerical\_columns(df, method='standard')} & Normalize all numerical columns using specified method \\
\hline
\texttt{concatenate\_train\_test(train\_df, test\_df)} & Concatenate train and test data with tracking columns for proper splitting \\
\hline
\texttt{split\_combined\_into\_train\_test(combined)} & Split combined data back into train and test using tracking columns \\
\hline
\texttt{convert\_dataframe\_to\_features\_target(df, target\_column, is\_train=True)} & Convert DataFrame to features and target format \\
\hline
\texttt{convert\_to\_dataframe(data, **kwargs)} & Convert various data types to pandas DataFrame \\
\hline
\texttt{drop\_feature(df, column)} & Drop feature(s) from the DataFrame \\
\hline
\texttt{get\_features(df, columns)} & Extract specific features (columns) from the DataFrame \\
\hline
\texttt{concatenate\_dataframes(df1, df2, axis=0)} & Concatenate two DataFrames \\
\hline
\texttt{join\_dataframes(left\_df, right\_df, left\_on, right\_on=None, how='inner', suffixes=('\_x', '\_y'))} & Join two DataFrames using pandas merge functionality \\
\hline
\texttt{rename\_feature(df, old\_name, new\_name)} & Rename feature(s) \\
\hline
\texttt{get\_unique\_values(df, column, sort=True, include\_counts=True)} & Get unique values from a column as a DataFrame \\
\hline
\texttt{get\_dataframe\_dtypes\_summary(df)} & Get comprehensive summary of the dtypes in the entire DataFrame \\
\hline
\texttt{filter\_dataframe(df, condition)} & Filter DataFrame using a boolean condition \\
\hline

\multicolumn{2}{|c|}{\textbf{Model Utilities}} \\
\hline
\texttt{save\_model(model, filepath='model.pkl')} & Save the trained model to disk using pickle \\
\hline
\texttt{load\_model(filepath)} & Load a trained model from disk using pickle \\
\hline
\texttt{save\_dataframe\_to\_csv(df, filepath)} & Save a DataFrame to CSV file \\
\hline

\multicolumn{2}{|c|}{\textbf{Model Evaluation Functions}} \\
\hline
\texttt{evaluate\_regression\_model(model, X\_test, y\_test, model\_name="model", eval\_data\_label='test')} & Evaluate a trained regression model on data \\
\hline
\texttt{evaluate\_classification\_model(model, X\_test, y\_test, model\_name="model", eval\_data\_label='test')} & Evaluate a trained classification model on data \\
\hline
\texttt{predict\_target(model, X\_data, model\_name="model", return\_probabilities=False)} & Make predictions using a trained model \\
\hline
\end{longtable}

\section{Tool Masking Ablations}
\label{ap:ablations}
In this section, we perform an ablation study to investigate if tool masking contributes significantly to the performance of Hierarchical MCTS. We select a subset of five Kaggle challenges from our benchmark and evaluate all of the planning approaches, alongside a Hierarchical MCTS approach that does not use tool masking, i.e all tools are available to the agent during all the subtasks. We use GPT-4.1-mini for our experiments instead of GPT-4o for cost reasons. The results are presented in Tables ~\ref{tab:consistency-ablation-gpt-4.1-mini} and ~\ref{tab:percentile-ablation-gpt-4.1-mini}. The results demonstrate that the performance of Hierarchical MCTS degrades substantially without tool masking. Hierarchical MCTS without tool masking, achieves a median consistency of 0.3 and a median leaderboard percentile position of 0 across the chosen subset of five challenges. In comparison, Hierarchical MCTS with tool masking achieves a median consistency and leaderboard percentile of 0.8 and 21.10 respectively. This highlights that both tool masking and subtask decomposition are critical for effectively solving long-horizon planning problems in high-dimensional action spaces using LLMs.

\begin{table}[!h]
\centering
\scriptsize
\caption{Consistency across five competitions for GPT-4.1-mini with six planning algorithm variants. Maximum values per row are highlighted. The results demonstrate that the performance of Hierarchical MCTS degrades substantially without tool masking.}
\label{tab:consistency-ablation-gpt-4.1-mini}
\begin{tabular}{p{3.0cm} p{1.2cm} p{1.2cm} p{1.2cm} p{1.2cm} p{1.2cm} p{1.2cm}}
\toprule
\textbf{Competition} & \textbf{ReAct} & \textbf{LATS-Reflection} & \textbf{MCTS-Outcome} & \textbf{MCTS-Shaped} & \textbf{Hierarchical MCTS} & \textbf{Hierarchical MCTS (No Tool Masking)} \\
\midrule
\href{https://www.kaggle.com/competitions/spaceship-titanic}{Spaceship Titanic} & 0.1 & 0.1 & 0.1 & \underline{\textbf{0.7}} & 0.5 & 0 \\
\href{https://www.kaggle.com/competitions/playground-series-s4e8}{Poisonous Mushroom Prediction} & 0.4 & 0.5 & 0.2 & \underline{\textbf{0.6}} & 0.3 & 0.2 \\
\href{https://www.kaggle.com/competitions/playground-series-s4e1}{Bank Churn Classification} & 0.3 & 0.3 & 0.3 & \underline{\textbf{0.9}} & \underline{\textbf{0.9}} & 0.3 \\
\href{https://www.kaggle.com/competitions/santander-customer-transaction-prediction}{Santander Customer Transaction Prediction} & 0.6 & 0.2 & 0.0 & 0.7 & \underline{\textbf{0.9}} & 0.3 \\
\href{https://www.kaggle.com/competitions/new-york-city-taxi-fare-prediction}{NYC Taxi Fare Prediction} & 0.0 & 0.0 & 0.4 & 0.1 & \underline{\textbf{0.8}} & 0.4 \\
\midrule
\textbf{Overall (Median)} & 0.3 & 0.2 & 0.2 & 0.7 & \underline{\textbf{0.8}} & 0.3 \\
\bottomrule
\end{tabular}
\end{table}

\begin{table}[!h]
\centering
\scriptsize
\caption{Median leaderboard percentile across five competitions for GPT-4.1-mini with six planning algorithm variants. Maximum values per row are highlighted. The results demonstrate that the performance of Hierarchical MCTS degrades substantially without tool masking.}
\label{tab:percentile-ablation-gpt-4.1-mini}
\begin{tabular}{p{3.0cm} p{1.2cm} p{1.2cm} p{1.2cm} p{1.2cm} p{1.2cm} p{1.2cm}}
\toprule
\textbf{Competition} & \textbf{ReAct} & \textbf{LATS-Reflection} & \textbf{MCTS-Outcome} & \textbf{MCTS-Shaped} & \textbf{Hierarchical MCTS} & \textbf{Hierarchical MCTS (No Tool Masking)} \\
\midrule
\href{https://www.kaggle.com/competitions/spaceship-titanic}{Spaceship Titanic} & 0.0 & 0.0 & 0.0 & \underline{\textbf{53.59}} & 28.01 & 0.0 \\
\href{https://www.kaggle.com/competitions/playground-series-s4e8}{Poisonous Mushroom Prediction} & 0.0 & 6.60 & 0.0 & \underline{\textbf{16.99}} & 0.0 & 0.0 \\
\href{https://www.kaggle.com/competitions/playground-series-s4e1}{Bank Churn Classification} & 0.0 & 0.0 & 0.0 & \underline{\textbf{31.57}} & \underline{\textbf{31.57}} & 0.0 \\
\href{https://www.kaggle.com/competitions/santander-customer-transaction-prediction}{Santander Customer Transaction Prediction} & \underline{\textbf{2.27}} & 0.0 & 0.0 & \underline{\textbf{2.27}} & \underline{\textbf{2.27}} & 0.0 \\
\href{https://www.kaggle.com/competitions/new-york-city-taxi-fare-prediction}{NYC Taxi Fare Prediction} & 0.0 & 0.0 & 0.0 & 0.0 & \underline{\textbf{21.10}} & 0.0 \\
\midrule
\textbf{Overall (Median)} & 0.0 & 0.0 & 0.0 & 16.99 & \underline{\textbf{21.10}} & 0.0 \\
\bottomrule
\end{tabular}
\end{table}

\section{Cost Comparisions}
\label{ap:cost}

\paragraph{Model Selection.} We evaluate GPT-4o and GPT-4.1-mini in our experiments. Frontier models such as GPT-5.2 and Claude Opus 4.5 were not included due to cost constraints. Relative to GPT-4.1-mini, GPT-5.2 costs approximately 4.4$\times$ more on input tokens and 8.8$\times$ more on output tokens, while Claude Opus 4.5 costs 12.5$\times$ more on input and 15.6$\times$ more on output. Given our evaluation protocol (10 trials per algorithm-challenge pair with hundreds of LLM calls per trajectory) and a total budget of \$5,000, evaluating these frontier models would have exceeded our budget.

\paragraph{Cost Comparison Across Methods.} We also provide cost comparisons for all planning algorithms using GPT-4.1-mini on the same subset of five Kaggle challenges from our benchmark that was used in the Tool Ablation Study (Appendix ~\ref{ap:ablations}). The results are reported in Table ~\ref{tab:cost-gpt-4.1-mini}. LATS is the most expensive planning approach, costing 3.5× more than the more successful variants (Hierarchical-MCTS and MCTS-Shaped), while achieving only a consistency of 0.2 and a median percentile position of 0 across the five Kaggle challenges. This suggests that LATS’s search is unfocused and tends to wander due to inconsistent scoring by the LLM evaluator. ReAct is the cheapest method but also performs poorly, only marginally outperforming LATS despite the latter using 10.5× more budget.

\begin{table}[!h]
\centering
\scriptsize
\caption{Total costs (\$) aggregated over 10 trajectories for each of the five competitions using GPT-4.1-mini under all planning algorithm variants examined in this study. For each row, the maximum value is highlighted}
\label{tab:cost-gpt-4.1-mini}
\begin{tabular}{p{3.0cm} p{1.2cm} p{1.2cm} p{1.2cm} p{1.2cm} p{1.2cm} p{1.2cm}}
\toprule
\textbf{Competition} & \textbf{ReAct} & \textbf{LATS-Reflection} & \textbf{MCTS-Outcome} & \textbf{MCTS-Shaped} & \textbf{Hierarchical MCTS} \\
\midrule
\href{https://www.kaggle.com/competitions/spaceship-titanic}{Spaceship Titanic} & 1.66 & \underline{\textbf{22.29}} & 3.22 & 8.44 & 3.93 \\
\href{https://www.kaggle.com/competitions/playground-series-s4e8}{Poisonous Mushroom Prediction} & 0.86 & \underline{\textbf{13.64}} & 3.1.3 & 2.86 & 1.89 \\
\href{https://www.kaggle.com/competitions/playground-series-s4e1}{Bank Churn Classification} & 1.21 & \underline{\textbf{9.98}} & 1.44 & 1.79 & 5.65 \\
\href{https://www.kaggle.com/competitions/santander-customer-transaction-prediction}{Santander Customer Transaction Prediction} & 1.34 & \underline{\textbf{11.28}} & 2.36 & 1.43 & 5.38 \\
\href{https://www.kaggle.com/competitions/new-york-city-taxi-fare-prediction}{NYC Taxi Fare Prediction} & 2.02 & \underline{\textbf{17.83}} & 2.72 & 4.07 & 5.56\\
\midrule
\textbf{Overall (Sum)} & 7.08 & \underline{\textbf{75.02}} & 12.86 & 18.59 & 22.42\\
\bottomrule
\end{tabular}
\end{table}

\section{Comparison with Direct Code Generation}\label{ap:compare-code-gen}

We compare our approaches with a direct code-generation variant, ReAct-Code, which produces end-to-end solutions for each challenge and iteratively refines code using execution feedback. The agent is allowed up to 10 iterations to refine its solution and produce a final submission file. We evaluate all algorithms on the original train and test splits of the Kaggle benchmark, consistent with Tables~\ref{tab:consistency-gpt-4.1-mini-kaggle-original} and~\ref{tab:leaderboard-percentile-gpt-4.1-mini-kaggle-original} in the main paper. Table~\ref{tab:consistency-react-code} presents Consistency scores across competitions, and Table~\ref{tab:leaderboard-react-code} shows the corresponding Kaggle public leaderboard percentile ranks.

\begin{table}[!h]
\centering
\scriptsize
\caption{Consistency across Kaggle competitions on the original train--test splits.}
\label{tab:consistency-react-code}
\begin{tabular}{lcccccc}
\toprule
\textbf{Competition} &
\textbf{ReAct} &
\textbf{LATS} &
\textbf{MCTS-Outcome} &
\textbf{MCTS-Shaped} &
\textbf{Hierarchical MCTS} &
\textbf{ReAct-Code} \\
\midrule
\href{https://www.kaggle.com/competitions/spaceship-titanic}{Spaceship Titanic} & 0.3 & 0.0 & 0.6 & 0.6 & 0.4 & 1.0 \\
\href{https://www.kaggle.com/competitions/playground-series-s5e9}{BPM Prediction}& 0.2 & 0.3 & 0.0 & 0.8 & 0.7 & 1.0 \\
\href{https://www.kaggle.com/competitions/playground-series-s5e5}{Calorie Expenditure Prediction} & 0.4 & 0.3 & 0.2 & 0.3 & 0.6 & 0.7 \\
\href{https://www.kaggle.com/competitions/playground-series-s3e1/overview}{California Housing Regression} & 0.6 & 0.4 & 0.1 & 0.5 & 0.8 & 0.8 \\
\href{https://www.kaggle.com/competitions/playground-series-s5e8}{Bank Deposit Classification} & 0.4 & 0.5 & 0.0 & 1.0 & 0.6 & 0.8 \\
\href{https://www.kaggle.com/competitions/playground-series-s4e1}{Bank Churn Classification} & 0.2 & 0.5 & 0.1 & 1.0 & 0.8 & 1.0 \\
\midrule
Overall (Median) & 0.35 & 0.35 & 0.1 & 0.7 & 0.65 & 0.9 \\
\bottomrule
\end{tabular}
\end{table}

\begin{table}[!h]
\centering
\scriptsize
\caption{Kaggle public leaderboard percentile across competitions.}
\label{tab:leaderboard-react-code}
\begin{tabular}{lcccccc}
\toprule
\textbf{Competition} &
\textbf{ReAct} &
\textbf{LATS} &
\textbf{MCTS-Outcome} &
\textbf{MCTS-Shaped} &
\textbf{Hierarchical MCTS} &
\textbf{ReAct-Code} \\
\midrule
\href{https://www.kaggle.com/competitions/spaceship-titanic}{Spaceship Titanic} & 0.00 & 0.00 & 39.77 & 41.69 & 0.00 & 48.08 \\
\href{https://www.kaggle.com/competitions/playground-series-s5e9}{BPM Prediction} & 0.00 & 0.00 & 0.00 & 5.03 & 0.19 & 6.75 \\
\href{https://www.kaggle.com/competitions/playground-series-s5e5}{Calorie Expenditure Prediction} & 0.00 & 0.00 & 0.00 & 0.00 & 16.81 & 24.29 \\
\href{https://www.kaggle.com/competitions/playground-series-s3e1/overview}{California Housing Regression} & 5.51 & 0.00 & 0.00 & 8.99 & 24.78 & 29.93 \\
\href{https://www.kaggle.com/competitions/playground-series-s5e8}{Bank Deposit Classification}& 0.00 & 13.47 & 0.00 & 27.65 & 27.29 & 37.48 \\
\href{https://www.kaggle.com/competitions/playground-series-s4e1}{Bank Churn Classification} & 0.00 & 14.91 & 0.00 & 29.87 & 32.47 & 35.09 \\
\midrule
Overall (Median) & 0.00 & 0.00 & 0.00 & 18.32 & 20.80 & 32.51 \\
\bottomrule
\end{tabular}
\end{table}

ReAct-Code, which directly generates executable code to solve the Kaggle challenges, outperforms all planning-oriented methods on both Consistency and leaderboard percentile. This result is expected: direct code generation has access to the full modeling and data-processing pipeline, whereas planning-based methods are constrained to a limited toolset and therefore face a strictly harder problem. Furthermore, LLMs may have been pretrained on Kaggle challenge solutions, whereas in planning-based variants the agent must explicitly discover a performant solution trajectory using the available tools. Importantly, despite these restrictions, our planning approaches do not lag too far behind ReAct-Code, suggesting that the chosen toolset is largely sufficient for solving tabular Kaggle ML tasks.

\section{Prompts}\label{ap:prompts}

For each competition in our benchmark, we constructed a standardized instruction template that included: (i) a brief description of the Kaggle challenge, (ii) a description of the data fields, and (iii) additional requirements specified from our end regarding model training and submission. For the challenge and data field descriptions, we used only the information provided in the corresponding sections of the original Kaggle competition; no human-authored modifications or additions were introduced.

Below, we show the exact template used for the \textit{\href{https://www.kaggle.com/competitions/spaceship-titanic}{Spaceship Titanic}} challenge \citep{spaceship-titanic}:

\begin{quote}
Welcome to the year 2912, where your data science skills are needed to solve a cosmic mystery. We've received a transmission from four lightyears away and things aren't looking good.

The Spaceship Titanic was an interstellar passenger liner launched a month ago. With almost 13{,}000 passengers on board, the vessel set out on its maiden voyage transporting emigrants from our solar system to three newly habitable exoplanets orbiting nearby stars.

While rounding Alpha Centauri en route to its first destination—the torrid 55 Cancri E—the unwary Spaceship Titanic collided with a spacetime anomaly hidden within a dust cloud. Sadly, it met a similar fate as its namesake from 1000 years before. Though the ship stayed intact, almost half of the passengers were transported to an alternate dimension! Read the data from the train.csv file in the data folder.

To help rescue crews and retrieve the lost passengers, you are challenged to predict which passengers were transported by the anomaly using records recovered from the spaceship's damaged computer system.

In this competition your task is to predict whether a passenger was transported to an alternate dimension during the Spaceship Titanic's collision with the spacetime anomaly. To help you make these predictions, you're given a set of personal records recovered from the ship's damaged computer system.

\textbf{File and Data Field Descriptions}
\begin{itemize}
    \item \texttt{PassengerId} -- Unique identifier in the form gggg\_pp, where gggg indicates a group and pp a member index.
    \item \texttt{HomePlanet} -- Planet of permanent residence.
    \item \texttt{CryoSleep} -- Whether the passenger elected suspended animation.
    \item \texttt{Cabin} -- Cabin number, formatted as deck/num/side.
    \item \texttt{Destination} -- Planet of disembarkation.
    \item \texttt{Age} -- Age of passenger.
    \item \texttt{VIP} -- Whether the passenger paid for VIP services.
    \item \texttt{RoomService, FoodCourt, ShoppingMall, Spa, VRDeck} -- Spending at onboard amenities.
    \item \texttt{Name} -- Full passenger name.
    \item \texttt{Transported} -- Target variable --- whether the passenger was transported.
\end{itemize}

\textbf{Submission File Format}
\begin{itemize}
    \item \texttt{PassengerId} -- Identifier for each passenger in the test set.
    \item \texttt{Transported} -- Prediction (True/False).
\end{itemize}

\textbf{Benchmark Instructions}
\begin{itemize}
    \item The training data is located at \texttt{data/spaceship\_titanic/train.csv}.
    \item The test data is located at \texttt{data/spaceship\_titanic/test.csv}.
    \item Load, clean, and perform feature engineering before fitting models.
    \item Concatenate train and test datasets before preprocessing to ensure consistent transformations, then split back.
    \item Experiment with multiple models and hyperparameter tuning to find the best-performing solution.
    \item Report evaluation results demonstrating model fit.
    \item Save the best model to \texttt{model\_saves/spaceship\_titanic/}.
    \item Save predictions for the test set in CSV format to \{\texttt{save\_directory}\}/\{\texttt{save\_file\_name}\}.csv.
\end{itemize}
\end{quote}

\subsection{LATS}

For LATS \citep{zhou2024languageagenttreesearch}, we used the following system prompt for the reflection step, where an LLM was asked to evaluate the quality of a given trajectory (i.e., all message contents of the trajectory were passed as input):

\begin{quote}
You are a Data Science judge, who evaluates the goodness of tool calling trajectories to solve Machine Learning tasks on Kaggle. Reflect and grade the agent's trajectory plan for the provided challenge. The trajectories should be aimed towards solving the challenge, i.e., generating a trained model and a valid submission file. Keep your reflections concise and to the point.
\end{quote}

For the expansion stage, we used the following system prompt. The trajectory of messages up to the current node was passed as input, and the agent was asked to propose new expansion candidates:

\begin{quote}
You are a Data Scientist tasked with solving the Kaggle competition provided below, with the tools available to you. Propose tool candidates that would help solve the problem at the current stage.
\end{quote}

\subsection{MCTS}

For both MCTS-Outcome and MCTS-Shaped, we used the same system prompt in the expansion stage, as was used for LATS. The trajectory of messages up to the current node was passed as input, and the agent was asked to propose new expansion candidates (same as LATS)

\subsection{Hierarchical MCTS}

For Hierarchical-MCTS, to propose candidates during the expansion phase, we used subtask-specific prefixes for the system prompt. Each subtask was associated with a descriptive prefix that constrained the role of the agent and defined the completion condition for that stage. The list of prefixes is shown below:

\begin{itemize}
    \item \texttt{train\_data\_loading}: ``You are a Data Scientist in the Data Loading stage of solving a Kaggle challenge, using only the tools available to you. This stage ends when you have loaded the train data successfully.''
    \item \texttt{test\_data\_loading}: ``You are a Data Scientist in the Data Loading stage of solving a Kaggle challenge, using only the tools available to you. This stage ends when you have loaded the test data successfully.''
    \item \texttt{combine\_train\_test}: ``You are a Data Scientist in the Data Loading stage of solving a Kaggle challenge, using only the tools available to you. This stage ends when you have combined the train and test data into a single dataframe successfully, to be used for downstream Data Cleaning and Feature Engineering.''
    \item \texttt{data\_cleaning}: ``You are a Data Scientist in the Data Cleaning stage of solving a Kaggle challenge, using only the tools available to you. This stage ends when there are no missing values present in the data. This also includes the column corresponding to the target variable, that may have NaNs in the test partition since the target variable is not present in the test partition. You are allowed to be innovative in filling the missing values based on your understanding of the data.''
    \item \texttt{feature\_engineering}: ``You are a Data Scientist in the Feature Engineering stage of solving a Kaggle challenge, using only the tools available to you. Create new features, or delete unimportant features or transform existing features as needed. You are not allowed to delete or modify features that indicate if the row in the data belongs to the train or test partition. You are also not allowed to augment the feature corresponding to the target variable. Use your understanding of the data to aid your decisions. This stage ends when the models feel that the features are good enough for modeling, and categorical and numerical features have been properly encoded. After the end of this stage, all the features should be (i) either int or float or (ii) int, float, category with the number of unique values in the category columns not being exorbitantly large.''
    \item \texttt{split\_train\_test}: ``You are a Data Scientist in the Split Train Test stage of solving a Kaggle challenge, using only the tools available to you. Split the combined train and test data into train and test dataframes. This stage ends when the train and test dataframes are successfully split from the combined dataframe.''
    \item \texttt{train\_data\_to\_features\_target}: ``You are a Data Scientist in the Converting the Train Data to Features and Target stage of solving a Kaggle challenge, using only the tools available to you. Convert the train data into features and target. This stage ends when the train data is successfully converted into features and target, for making downstream modeling upon.''
    \item \texttt{test\_data\_to\_features}: ``You are a Data Scientist in the Converting the Test Data to Features stage of solving a Kaggle challenge, using only the tools available to you. Convert the test data into features. This stage ends when the test data is successfully converted into features, for making downstream predictions upon.''
    \item \texttt{modeling}: ``You are a Data Scientist in the Modeling stage of solving a Kaggle challenge, using only the tools available to you. Train and tune models. You might need to experiment with different model choices and properly tune your hyperparameters to get good performance. Use the provided evaluation tools to evaluate your trained models if needed. This stage ends when the agent has successfully created a model that it considers to be the best.''
    \item \texttt{create\_submission\_dataframe}: ``You are a Data Scientist in the Create Submission stage of solving a Kaggle challenge, using only the tools available to you. Make predictions on the test data, and create a submission dataframe that contains the predictions in the requested format. This stage ends when the submission dataframe in the correct format is created.''
\end{itemize}

The system prompt was then constructed as:
\begin{quote}
\{\texttt{subtask\_description}\}. Propose tool candidates that would help solve the problem at the current stage.
\end{quote}

The trajectory of messages up to the current node was passed as input, and the agent was asked to propose new expansion candidates (same as what was used for LATS, MCTS variants)

\section{Visualizing Trajectories}
\label{ap:visualize}
In this section, we present example trajectories for each of the planning algorithms evaluated in this paper. For every algorithm, we provide both a successful and a failed trajectory. All trajectories are collected on the Spaceship Titanic Kaggle challenge using the GPT-4.1-mini model, ensuring a consistent evaluation setting across planning methods.

\begin{table}[t]
    \centering
    \caption{Comparison of tool usage and performance across planning algorithms for GPT-4.1-mini on the Spaceship Titanic challenge. The tool calls columns report the number of tool calls made by the agent in failed and successful trajectories, respectively. The successful trajectory performance denotes the leaderboard percentile position achieved by the submission file generated from the successful trajectory.}
    \label{tab:tool_calls_case_study}
    \begin{tabular}{p{3.4cm} p{3.2cm} p{3.4cm} p{3.4cm}}
        \toprule
        \textbf{Algorithm} &
        \textbf{Failed Trajectory Tool Calls} &
        \textbf{Successful Trajectory Tool Calls} &
        \textbf{Successful Trajectory Performance} \\
        \midrule
        ReAct             & 12 & 40 & 54.92 \\
        LATS              & 17 & 27 & 58.21 \\
        MCTS-Outcome      & 18 & 42 & 0.00  \\
        MCTS-Shaped       & 19 & 23 & 33.11 \\
        Hierarchical-MCTS & 55 & 80 & 62.65 \\
        \bottomrule
    \end{tabular}
\end{table}

In Table ~\ref{tab:tool_calls_case_study} above, the tool calls columns denote the number of tool calls made by the agent in the corresponding trajectory. The successful trajectory performance denotes the leaderboard percentile position of the submission file created by the successful trajectory for each planning algorithm.

Across all algorithms except Hierarchical MCTS, only a single valid model was created before the agent transitioned to the evaluation phase. In contrast, Hierarchical MCTS experimented with multiple machine learning models, including Random Forest, LightGBM, and CatBoost, and selected the best-performing model. We note that the failed trajectory in LATS did fit a model; however, the LLM evaluator preferred other suboptimal trajectories, highlighting potential issues in LLM-based state estimation. Failed trajectories in ReAct also exhibited premature termination before task completion. Additionally, some failed trajectories were unable to invoke tools successfully or struggled to reach the modeling phase due to insufficient guidance, as observed in MCTS-Outcome.

We further note that the successful trajectory for MCTS-Outcome in this case study did produce a submission file; however, it exhibited a shape mismatch with the original test data and was consequently assigned a leaderboard percentile position of 0. In contrast, the successful trajectory of Hierarchical MCTS used 80 tool calls, explored multiple modeling strategies, and achieved a leaderboard percentile position of 62.65.

\subsection{ReAct}

\subsubsection{Failed Trajectory}

The following failed ReAct trajectory contains 12 tool calls and terminates during the Feature Engineering stage, never reaching the Modeling stage. This pattern is common among failed ReAct runs: a substantial fraction are unable to complete the full end-to-end machine learning pipeline required for the Kaggle task. This snippet corresponds to the final portion of the failed ReAct trajectory.

\begin{lstlisting}[basicstyle=\ttfamily\small]
{
  "step_number": 20,
  "timestamp": "2025-11-20T18:52:06.565366",
  "step_type": "tool_execution_initiation",
  "action": "tool_execution_started",
  "tools_to_execute": [
    {
      "tool_name": "apply_custom_function",
      "tool_args": {
        "bindings": { "df": "combined_data" },
        "output": "combined_data",
        "func_kwargs": {
          "name": "Cabin_deck",
          "source_columns": "Cabin",
          "func": "lambda x: x.split('/')[0] if isinstance(x, str) else None"
        }
      },
      "tool_id": "call_s5ZjVgGT9j6llr3Vg7jUQavA"
    }
  ]
},
{
  "step_number": 21,
  "timestamp": "2025-11-20T18:52:06.599497",
  "step_type": "tool_execution_completion",
  "action": "tool_execution_completed",
  "tool_results": [
    {
      "message_type": "ToolMessage",
      "content_preview": "Error: TypeError(\"'str' object is not callable\")\n Please fix your mistakes.",
      "content_length": 75
    }
  ]
},
{
  "step_number": 22,
  "timestamp": "2025-11-20T18:52:10.291750",
  "step_type": "tool_execution_initiation",
  "action": "tool_execution_started",
  "tools_to_execute": [
    {
      "tool_name": "apply_custom_function",
      "tool_args": {
        "bindings": { "df": "combined_data" },
        "output": "combined_data",
        "func_kwargs": {
          "name": "Cabin_deck",
          "source_columns": "Cabin",
          "func": null
        }
      },
      "tool_id": "call_JiDjhYHW16xgodP4y69hOAR3"
    }
  ]
},
{
  "step_number": 23,
  "timestamp": "2025-11-20T18:52:10.324847",
  "step_type": "tool_execution_completion",
  "action": "tool_execution_completed",
  "tool_results": [
    {
      "message_type": "ToolMessage",
      "content_preview": "Error: TypeError(\"'NoneType' object is not callable\")\n Please fix your mistakes.",
      "content_length": 80
    }
  ]
}
\end{lstlisting}

\subsubsection{Successful Trajectory}

This trajectory comprises 40 tool calls and demonstrates strong performance. The feature engineering stage involves numerous successful and relevant tool invocations. In contrast, the modeling stage fits only a CatBoost classifier without exploring additional model families. When scored on our internal test set, this trajectory attains a percentile of 54.92 on the Kaggle leaderboard for this task.

\begin{lstlisting}
    {
      "step_number": 70,
      "timestamp": "2025-11-20T18:53:09.461595",
      "step_type": "tool_execution_initiation",
      "action": "tool_execution_started",
      "tools_to_execute": [
        {
          "tool_name": "convert_dataframe_to_features_target",
          "tool_args": {
            "bindings": {
              "df": "test_df"
            },
            "func_kwargs": {
              "target_column": "Transported",
              "is_train": false
            },
            "output": "test_features_target"
          },
          "tool_id": "call_VTa2DIpTuPJRLydhruqaaiBK"
        }
      ]
    },
    {
      "step_number": 71,
      "timestamp": "2025-11-20T18:53:09.535408",
      "step_type": "tool_execution_completion",
      "action": "tool_execution_completed",
      "tool_results": [
        {
          "message_type": "ToolMessage",
          "content_preview": "Applied convert_dataframe_to_features..
          "content_length": 864
        }
      ]
    },
    {
      "step_number": 72,
      "timestamp": "2025-11-20T18:53:11.188346",
      "step_type": "tool_execution_initiation",
      "action": "tool_execution_started",
      "tools_to_execute": [
        {
          "tool_name": "tune_catboost_classifier",
          "tool_args": {
            "bindings": {
              "X_train": "X_train",
              "y_train": "Y_train"
            },
            "output": "catboost_tuning_result"
          },
          "tool_id": "call_DHeRtQVO4RZb8ayZyc47Ow0t"
        }
      ]
    },
    {
      "step_number": 73,
      "timestamp": "2025-11-20T18:53:12.310197",
      "step_type": "tool_execution_completion",
      "action": "tool_execution_completed",
      "tool_results": [
        {
          "message_type": "ToolMessage",
          "content_preview": "Applied tune_catboost_classifier. The Best params and CV score for this method are {'depth': 6, 'iterations': 50, 'l2_leaf_reg': 1, 'learning_rate': 0.1} and 0.8011213918872091 respectively. The mapping between the function parameters and the scratchpad k... [truncated]",
          "content_length": 1088
        }
      ]
    },
    {
      "step_number": 74,
      "timestamp": "2025-11-20T18:53:14.269490",
      "step_type": "tool_execution_initiation",
      "action": "tool_execution_started",
      "tools_to_execute": [
        {
          "tool_name": "evaluate_classification_model",
          "tool_args": {
            "bindings": {
              "model": "best_estimator",
              "X_test": "X_train",
              "y_test": "Y_train"
            },
            "func_kwargs": {
              "model_name": "CatBoost Classifier",
              "eval_data_label": "train"
            }
          },
          "tool_id": "call_76EXOjjGyUp86wMwe3bVbES4"
        }
      ]
    },
    {
      "step_number": 75,
      "timestamp": "2025-11-20T18:53:14.363633",
      "step_type": "tool_execution_completion",
      "action": "tool_execution_completed",
      "tool_results": [
        {
          "message_type": "ToolMessage",
          "content_preview": "Applied evaluate_classification_model.. [truncated]",
          "content_length": 1753
        }
      ]
    },
    {
      "step_number": 76,
      "timestamp": "2025-11-20T18:53:16.071819",
      "step_type": "tool_execution_initiation",
      "action": "tool_execution_started",
      "tools_to_execute": [
        {
          "tool_name": "predict_target",
          "tool_args": {
            "bindings": {
              "model": "best_estimator",
              "X_data": "X_test"
            },
            "func_kwargs": {
              "model_name": "CatBoost Classifier"
            },
            "output": "test_predictions"
          },
          "tool_id": "call_fM4PS3vRAthZWoDZw6ycGzUQ"
        }
      ]
    },
    {
      "step_number": 77,
      "timestamp": "2025-11-20T18:53:16.166858",
      "step_type": "tool_execution_completion",
      "action": "tool_execution_completed",
      "tool_results": [
        {
          "message_type": "ToolMessage",
          "content_preview": "Applied predict_target ..[truncated]",
          "content_length": 1019
        }
      ]
    },
    {
      "step_number": 78,
      "timestamp": "2025-11-20T18:53:26.236014",
      "step_type": "tool_execution_initiation",
      "action": "tool_execution_started",
      "tools_to_execute": [
        {
          "tool_name": "save_model",
          "tool_args": {
            "bindings": {
              "model": "best_estimator"
            },
            "func_kwargs": {
              "filepath": "model_saves/spaceship_titanic/catboost_best_model.pkl"
            }
          },
          "tool_id": "call_PI0hV7OnLQEBSrir8PNgWSU0"
        }
      ]
    },
    {
      "step_number": 79,
      "timestamp": "2025-11-20T18:53:26.322763",
      "step_type": "tool_execution_completion",
      "action": "tool_execution_completed",
      "tool_results": [
        {
          "message_type": "ToolMessage",
          "content_preview": "Applied save_model ..",
          "content_length": 676
        }
      ]
    },
    {
      "step_number": 80,
      "timestamp": "2025-11-20T18:53:29.230047",
      "step_type": "tool_execution_initiation",
      "action": "tool_execution_started",
      "tools_to_execute": [
        {
          "tool_name": "save_dataframe_to_csv",
          "tool_args": {
            "bindings": {
              "df": "test_predictions"
            },
            "func_kwargs": {
              "filepath": "/submissions/spaceship_titanic/react/react_agent_4_47a197f8-edb6-4c43-a9ce-ae5f8ef45736.csv"
            }
          },
          "tool_id": "call_xrfZg9Pg0FMXxgpr7a2rAGxf"
        }
      ]
    },
    {
      "step_number": 81,
      "timestamp": "2025-11-20T18:53:29.329762",
      "step_type": "tool_execution_completion",
      "action": "tool_execution_completed",
      "tool_results": [
        {
          "message_type": "ToolMessage",
          "content_preview": "Applied save_dataframe_to_csv..",
          "content_length": 777
        }
      ]
    },
    {
      "step_number": 82,
      "timestamp": "2025-11-20T18:53:34.098547",
      "step_type": "execution_summary",
      "action": "agent_execution_completed",
      "total_execution_time": 114.67485404014587,
      "total_tokens": 1022157,
      "total_cost": 0.156462,
      "final_message_count": 81,
      "competition_name": "spaceship_titanic"
    }
  ]
}
\end{lstlisting}

\subsection{LATS}

\subsection{Failed Trajectory}

This trajectory contains 17 tool calls. Although this trajectory, which denotes the agent’s most promising branch,  successfully fit a model, LATS ultimately prioritized other failed branches, causing it to miss this viable solution path. As a result, the final output did not include any valid trajectory that solved the challenge. This failure stems from inconsistencies in the reflection-based rating process, which incorrectly ranked failed nodes above successful progress.

\begin{lstlisting}
    {
      "step_number": 37,
      "message_type": "AIMessage",
      "timestamp": "2025-11-21T00:42:26.668798",
      "step_type": "tool_selection",
      "action": "selected_tools_for_execution",
      "tools_selected": [
        "tune_lightgbm_classifier"
      ],
      "tool_calls_detail": [
        {
          "name": "tune_lightgbm_classifier",
          "args": {
            "bindings": {
              "X_train": "X_train",
              "y_train": "Y_train"
            },
            "output": "best_lgbm_model"
          },
          "id": "call_iff8Jho9Tc91a1v1bEPfxIsZ",
          "type": "tool_call"
        }
      ],
      "content": "Features and target variables are now separated. I will proceed with hyperparameter tuning for a classification model to find the best model for this task. I will start with a LightGBM Classifier tuned using cross-validation."
    },
    {
      "step_number": 38,
      "message_type": "ToolMessage",
      "timestamp": "2025-11-21T00:42:26.668802",
      "step_type": "tool_result",
      "action": "received_tool_output",
      "content_preview": "Error: ValueError('Input y contains NaN.')\n Please fix your mistakes.",
      "content_length": 69
    },
    {
      "step_number": 39,
      "message_type": "HumanMessage",
      "timestamp": "2025-11-21T00:42:26.668807",
      "step_type": "reflection",
      "action": "llm_reflection",
      "content_preview": "Reasoning: The agent failed to check for and handle missing target values before model tuning, resulting in a ValueError due to NaNs in the target variable. This indicates a gap in data cleaning and validation steps before model training. Proper imputation or removal of missing target rows is necess...",
      "content_length": 704,
      "extracted_score": 4.0,
      "full_reflection_content": "Reasoning: The agent failed to check for and handle missing target values before model tuning, resulting in a ValueError due to NaNs in the target variable. This indicates a gap in data cleaning and validation steps before model training. Proper imputation or removal of missing target rows is necessary. The agent should have performed or verified missing value handling comprehensively before tuning models. This oversight blocks progress in model fitting and prediction generation, which are critical to solving the task. The trajectory is incomplete and insufficient as no valid model training was achieved due to missing data issues. Addressing this gap is essential for a correct solution.\nScore: 4"
    },
    {
      "step_number": 40,
      "message_type": "AIMessage",
      "timestamp": "2025-11-21T00:42:26.668821",
      "step_type": "tool_selection",
      "action": "selected_tools_for_execution",
      "tools_selected": [
        "filter_dataframe"
      ],
      "tool_calls_detail": [
        {
          "name": "filter_dataframe",
          "args": {
            "bindings": {
              "df": "combined_df"
            },
            "output": "combined_train_df",
            "func_kwargs": {
              "condition": "Transported.notna()"
            }
          },
          "id": "call_seR8LUmqkK389bcayn0QbxWI",
          "type": "tool_call"
        }
      ],
      "content": "The model tuning failed because there are missing values in the target column \"Transported\". I need to clean the data by removing rows with missing target values before proceeding with model training. I will filter out such rows from the combined dataset and then separate features and target again."
    },
    {
      "step_number": 41,
      "message_type": "ToolMessage",
      "timestamp": "2025-11-21T00:42:26.668824",
      "step_type": "tool_result",
      "action": "received_tool_output",
      "content_preview": "Applied filter_dataframe with docstring: \n    Filter DataFrame using a boolean condition.\n    \n    Parameters:\n    -----------\n    df : pd.DataFrame\n        Input DataFrame\n    condition : str\n        Boolean condition expression (e.g., 'col1 > 0', 'col1 == \"value\"', 'col1 > 0 and col2 < 100')\n    \n...",
      "content_length": 1141
    },
    {
      "step_number": 42,
      "message_type": "HumanMessage",
      "timestamp": "2025-11-21T00:42:26.668830",
      "step_type": "reflection",
      "action": "llm_reflection",
      "content_preview": "Reasoning: The agent initially struggled with the correct function call syntax for loading and manipulating data, causing delays in early steps. It successfully loaded and concatenated the train and test datasets, summarized missing values, cast categorical columns, and separated features and target...",
      "content_length": 850,
      "extracted_score": 5.0,
      "full_reflection_content": "Reasoning: The agent initially struggled with the correct function call syntax for loading and manipulating data, causing delays in early steps. It successfully loaded and concatenated the train and test datasets, summarized missing values, cast categorical columns, and separated features and target variables. However, the agent overlooked checking and handling missing target values before model training, leading to a model fitting failure. Once identified, it correctly filtered out rows with missing targets to fix this issue. The trajectory shows gradual improvement and understanding but is currently incomplete as it has not proceeded to feature engineering, model training, evaluation, or prediction saving yet. The agent needs to continue systematically through these steps to fully solve the task.\n\nScore: 5\nFound_solution: False\nScore: 5"
    },
    {
      "step_number": 43,
      "message_type": "AIMessage",
      "timestamp": "2025-11-21T00:42:26.668845",
      "step_type": "tool_selection",
      "action": "selected_tools_for_execution",
      "tools_selected": [
        "convert_dataframe_to_features_target"
      ],
      "tool_calls_detail": [
        {
          "name": "convert_dataframe_to_features_target",
          "args": {
            "bindings": {
              "df": "combined_train_df"
            },
            "output": "features_target_tuple_cleaned",
            "func_kwargs": {
              "target_column": "Transported",
              "is_train": true
            }
          },
          "id": "call_rm2jT91kE90EoOciGAwUXnco",
          "type": "tool_call"
        }
      ],
      "content": ""
    },
    {
      "step_number": 44,
      "message_type": "ToolMessage",
      "timestamp": "2025-11-21T00:42:26.668849",
      "step_type": "tool_result",
      "action": "received_tool_output",
      "content_preview": "Applied convert_dataframe_to_features_target with docstring: \n    Convert DataFrame to features and target format.\n    \n    Parameters:\n    -----------\n    df : pd.DataFrame\n        Input DataFrame\n    target_column : str\n        Name of the target column\n    is_train : bool\n        Whether the Data...",
      "content_length": 888
    },
    {
      "step_number": 45,
      "message_type": "HumanMessage",
      "timestamp": "2025-11-21T00:42:26.668854",
      "step_type": "reflection",
      "action": "llm_reflection",
      "content_preview": "Reasoning: The trajectory initially suffered from repeated syntax errors in function calls, blocking early progress on data loading. After correcting these, the agent effectively loaded train and test data, summarized missing values, concatenated datasets, and cast categorical columns properly. A ke...",
      "content_length": 949,
      "extracted_score": 6.0,
      "full_reflection_content": "Reasoning: The trajectory initially suffered from repeated syntax errors in function calls, blocking early progress on data loading. After correcting these, the agent effectively loaded train and test data, summarized missing values, concatenated datasets, and cast categorical columns properly. A key oversight was the presence of missing target values, which caused model tuning to fail. The agent correctly identified this and filtered out rows with missing targets before proceeding. The agent has now prepared clean training data and is ready for model tuning. However, feature engineering and model evaluation steps are still pending. The trajectory shows a solid recovery from early mistakes but remains incomplete in solving the full task of generating a trained model and valid predictions. Continued systematic progress through cleaning, feature engineering, modeling, and prediction saving is needed to fully solve the challenge.\nScore: 6"
    },
    {
      "step_number": 46,
      "message_type": "AIMessage",
      "timestamp": "2025-11-21T00:42:26.668870",
      "step_type": "tool_selection",
      "action": "selected_tools_for_execution",
      "tools_selected": [
        "tune_lightgbm_classifier"
      ],
      "tool_calls_detail": [
        {
          "name": "tune_lightgbm_classifier",
          "args": {
            "bindings": {
              "X_train": "X_train",
              "y_train": "Y_train"
            },
            "output": "best_lgbm_model"
          },
          "id": "call_SraYlT5IClOasjkTW9oju8xt",
          "type": "tool_call"
        }
      ],
      "content": ""
    },
    {
      "step_number": 47,
      "message_type": "ToolMessage",
      "timestamp": "2025-11-21T00:42:26.668873",
      "step_type": "tool_result",
      "action": "received_tool_output",
      "content_preview": "Applied tune_lightgbm_classifier with docstring: \n    Perform hyperparameter tuning for LightGBM Classifier using GridSearchCV.\n\n    Parameters\n    ----------\n    X_train : pd.DataFrame\n        Training features\n    y_train : pd.Series\n        Training target\n    cv : int, default=5\n        Number o...",
      "content_length": 1092
    },
    {
      "step_number": 48,
      "message_type": "HumanMessage",
      "timestamp": "2025-11-21T00:42:26.668878",
      "step_type": "reflection",
      "action": "llm_reflection",
      "content_preview": "Reasoning: The agent initially struggled with the correct syntax for function calls, which delayed progress in loading and concatenating data. However, the agent quickly recovered by successfully loading both train and test datasets, summarizing missing values, concatenating datasets, and correctly ...",
      "content_length": 1163,
      "extracted_score": 6.0,
      "full_reflection_content": "Reasoning: The agent initially struggled with the correct syntax for function calls, which delayed progress in loading and concatenating data. However, the agent quickly recovered by successfully loading both train and test datasets, summarizing missing values, concatenating datasets, and correctly casting categorical columns. A notable oversight was the failure to check for missing target values before model tuning, which resulted in an error. The agent fixed this by filtering out rows with missing targets. Subsequently, the agent performed hyperparameter tuning for a LightGBM classifier and obtained a best CV score of about 0.795, which is a good baseline for the classification task. The agent has established a good foundation but has not clearly shown subsequent steps such as evaluation on a validation set, feature engineering enhancements, saving the best model, or generating and saving predictions for the test set submission file. To fully solve the challenge, these remaining steps are needed. Overall, the trajectory shows reasonable understanding and recovery from initial mistakes but remains incomplete in finalizing the solution.\nScore: 6"
    },

\end{lstlisting}

\subsubsection{Successful Trajectory}

This successful trajectory comprises 27 tool calls and attains a leaderboard percentile of 58.21 on our test split. It is worth noting that LATS exclusively employed the LightGBM classifier and did not consider other modeling alternatives. 

\begin{lstlisting}
    {
      "step_number": 61,
      "message_type": "AIMessage",
      "timestamp": "2025-11-20T23:52:58.042763",
      "step_type": "tool_selection",
      "action": "selected_tools_for_execution",
      "tools_selected": [
        "convert_dataframe_to_features_target"
      ],
      "tool_calls_detail": [
        {
          "name": "convert_dataframe_to_features_target",
          "args": {
            "bindings": {
              "df": "test_df"
            },
            "func_kwargs": {
              "target_column": "Transported",
              "is_train": false
            },
            "output": "X_test"
          },
          "id": "call_LNz1Q2BWh1hYxqkPIbk29ojD",
          "type": "tool_call"
        }
      ],
      "content": ""
    },
    {
      "step_number": 62,
      "message_type": "ToolMessage",
      "timestamp": "2025-11-20T23:52:58.042766",
      "step_type": "tool_result",
      "action": "received_tool_output",
      "content_preview": "Applied convert_dataframe_to_features_target with docstring: \n    Convert DataFrame to features and target format.\n    \n    Parameters:\n    -----------\n    df : pd.DataFrame\n        Input DataFrame\n    target_column : str\n        Name of the target column\n    is_train : bool\n        Whether the Data...",
      "content_length": 863
    },
    {
      "step_number": 63,
      "message_type": "HumanMessage",
      "timestamp": "2025-11-20T23:52:58.042770",
      "step_type": "reflection",
      "action": "llm_reflection",
      "content_preview": "Reasoning: The agent followed a generally appropriate and methodical trajectory, starting with loading and concatenating train and test datasets to allow consistent preprocessing. They identified and handled missing values using sensible imputation methods for categorical and numerical features. The...",
      "content_length": 1404,
      "extracted_score": 6.0,
      "full_reflection_content": "Reasoning: The agent followed a generally appropriate and methodical trajectory, starting with loading and concatenating train and test datasets to allow consistent preprocessing. They identified and handled missing values using sensible imputation methods for categorical and numerical features. The agent cast categorical columns correctly and performed meaningful feature engineering, including extracting cabin features, aggregating spending features, and computing group features from PassengerId, demonstrating good domain understanding.\n\nHowever, the agent made some critical mistakes in the execution, such as not splitting the combined dataset back into train and test before model training, causing NaNs in the target labels and failing model tuning initially. They corrected this splitting issue later but also had minor syntax errors in tool usage. The agent has yet to demonstrate the crucial steps of fitting models, hyperparameter tuning, evaluation, saving the best model, and generating and saving predictions in the required submission format and location.\n\nOverall, the trajectory is solid for data loading, cleaning, and feature engineering but incomplete for the full challenge solution. The agent needs to proceed with modeling and output saving to fully solve the task. The response quality is good for preliminary steps but insufficient for a complete solution.\n\nScore: 6\nScore: 6"
    },
    {
      "step_number": 64,
      "message_type": "AIMessage",
      "timestamp": "2025-11-20T23:52:58.042793",
      "step_type": "tool_selection",
      "action": "selected_tools_for_execution",
      "tools_selected": [
        "tune_lightgbm_classifier"
      ],
      "tool_calls_detail": [
        {
          "name": "tune_lightgbm_classifier",
          "args": {
            "bindings": {
              "X_train": "X_train",
              "y_train": "Y_train"
            },
            "output": "best_lgbm_model"
          },
          "id": "call_WZihuqBOZNl9QcPJNoH1PhVo",
          "type": "tool_call"
        }
      ],
      "content": ""
    },
    {
      "step_number": 65,
      "message_type": "ToolMessage",
      "timestamp": "2025-11-20T23:52:58.042796",
      "step_type": "tool_result",
      "action": "received_tool_output",
      "content_preview": "Applied tune_lightgbm_classifier with docstring: \n    Perform hyperparameter tuning for LightGBM Classifier using GridSearchCV.\n\n    Parameters\n    ----------\n    X_train : pd.DataFrame\n        Training features\n    y_train : pd.Series\n        Training target\n    cv : int, default=5\n        Number o...",
      "content_length": 1092
    },
    {
      "step_number": 66,
      "message_type": "HumanMessage",
      "timestamp": "2025-11-20T23:52:58.042802",
      "step_type": "reflection",
      "action": "llm_reflection",
      "content_preview": "Reasoning: The agent has followed a generally sound and methodical approach suitable for the Spaceship Titanic classification task. They correctly loaded and concatenated the datasets, handled missing values with sensible imputation, cast categorical variables, and performed meaningful feature engin...",
      "content_length": 1162,
      "extracted_score": 7.0,
      "full_reflection_content": "Reasoning: The agent has followed a generally sound and methodical approach suitable for the Spaceship Titanic classification task. They correctly loaded and concatenated the datasets, handled missing values with sensible imputation, cast categorical variables, and performed meaningful feature engineering such as extracting cabin features and group-related variables. The agent made some mistakes in tool usage and workflow order, notably not splitting combined data before modeling, causing initial tuning errors due to NaNs in the target. These were corrected later. The agent successfully performed hyperparameter tuning on a LightGBM classifier and obtained a reasonable CV score (~0.79), indicating a decent model fit. However, the agent has not yet shown the steps to save the best model or generate and save predictions on the test set as required. The trajectory is strong in data preparation and modeling but incomplete on final output generation and saving, which is essential for solving the task fully. Overall, the response is effective but partial, scoring high on foundational steps but lacking completion of the full solution pipeline.\nScore: 7"
    },
    {
      "step_number": 67,
      "message_type": "AIMessage",
      "timestamp": "2025-11-20T23:52:58.042822",
      "step_type": "tool_selection",
      "action": "selected_tools_for_execution",
      "tools_selected": [
        "evaluate_classification_model"
      ],
      "tool_calls_detail": [
        {
          "name": "evaluate_classification_model",
          "args": {
            "bindings": {
              "model": "best_estimator",
              "X_test": "X_train",
              "y_test": "Y_train"
            },
            "func_kwargs": {
              "model_name": "LightGBM",
              "eval_data_label": "train"
            }
          },
          "id": "call_18GzPmWkNs6zNLuY9s0774Cm",
          "type": "tool_call"
        }
      ],
      "content": ""
    },
    {
      "step_number": 68,
      "message_type": "ToolMessage",
      "timestamp": "2025-11-20T23:52:58.042825",
      "step_type": "tool_result",
      "action": "received_tool_output",
      "content_preview": "Applied evaluate_classification_model with docstring: \n    Evaluate a trained classification model on data.\n\n    This function is used to evaluate the performance of a trained classification model on a given dataset.\n\n    The dataset can be the anything, such as the test set or the train set.\n\n    U...",
      "content_length": 1820
    },
    {
      "step_number": 69,
      "message_type": "HumanMessage",
      "timestamp": "2025-11-20T23:52:58.042829",
      "step_type": "reflection",
      "action": "llm_reflection",
      "content_preview": "Reasoning: The agent followed a standard and appropriate pipeline for the Spaceship Titanic classification task by loading and concatenating train and test datasets, imputing missing values, casting categorical variables, and executing meaningful feature engineering (extracting deck and side from Ca...",
      "content_length": 1303,
      "extracted_score": 7.0,
      "full_reflection_content": "Reasoning: The agent followed a standard and appropriate pipeline for the Spaceship Titanic classification task by loading and concatenating train and test datasets, imputing missing values, casting categorical variables, and executing meaningful feature engineering (extracting deck and side from Cabin, aggregating spending, and creating group features). They correctly split combined data back to train and test before modeling, which fixed an earlier critical workflow mistake. The agent performed hyperparameter tuning on a LightGBM classifier and achieved a reasonable CV score (~0.79) and training accuracy (~81%
    },
    {
      "step_number": 70,
      "message_type": "AIMessage",
      "timestamp": "2025-11-20T23:52:58.042851",
      "step_type": "tool_selection",
      "action": "selected_tools_for_execution",
      "tools_selected": [
        "predict_target"
      ],
      "tool_calls_detail": [
        {
          "name": "predict_target",
          "args": {
            "bindings": {
              "model": "best_estimator",
              "X_data": "X_test"
            },
            "func_kwargs": {
              "model_name": "LightGBM",
              "return_probabilities": false
            },
            "output": "test_predictions"
          },
          "id": "call_9j7gLB7G1cH6WGnnRpygvWVb",
          "type": "tool_call"
        }
      ],
      "content": ""
    },
    {
      "step_number": 71,
      "message_type": "ToolMessage",
      "timestamp": "2025-11-20T23:52:58.042854",
      "step_type": "tool_result",
      "action": "received_tool_output",
      "content_preview": "Applied predict_target with docstring: \n    Make predictions using a trained model.\n    \n    Parameters\n    ----------\n    model : Any\n        Trained model (regression or classification)\n    X_data : pd.DataFrame\n        Features to make predictions on\n    model_name : str, default=\"model\"\n        ...",
      "content_length": 1018
    },
    {
      "step_number": 72,
      "message_type": "HumanMessage",
      "timestamp": "2025-11-20T23:52:58.042858",
      "step_type": "reflection",
      "action": "llm_reflection",
      "content_preview": "Reasoning: The agent followed a comprehensive and well-structured approach to the Spaceship Titanic classification task. They started by loading and concatenating the train and test datasets to ensure consistent preprocessing. They successfully identified missing data and applied appropriate imputat...",
      "content_length": 1588,
      "extracted_score": 7.0,
      "full_reflection_content": "Reasoning: The agent followed a comprehensive and well-structured approach to the Spaceship Titanic classification task. They started by loading and concatenating the train and test datasets to ensure consistent preprocessing. They successfully identified missing data and applied appropriate imputation strategies for categorical and numerical features. Categorical columns were correctly cast, and meaningful feature engineering was performed, including extracting cabin deck and side, aggregating spending features, and creating group features from PassengerId, demonstrating good domain understanding. The agent then corrected a critical oversight by properly splitting the combined dataset back into train and test before modeling, which allowed successful hyperparameter tuning of a LightGBM model, yielding a reasonable cross-validation score (~0.79) and training accuracy (~81%
    },
    {
      "step_number": 73,
      "message_type": "AIMessage",
      "timestamp": "2025-11-20T23:52:58.042883",
      "step_type": "tool_selection",
      "action": "selected_tools_for_execution",
      "tools_selected": [
        "save_dataframe_to_csv"
      ],
      "tool_calls_detail": [
        {
          "name": "save_dataframe_to_csv",
          "args": {
            "df": "test_predictions",
            "filepath": "submissions/spaceship_titanic/lats_reflection/lats_reflection.csv"
          },
          "id": "call_BTgHyO2TJNUOWcyx4YTiNwIr",
          "type": "tool_call"
        }
      ],
      "content": ""
    },


\end{lstlisting}

\subsection{MCTS-Outcome}

\subsubsection{Failed Trajectory}

The most promising trajectory identified consisted of 18 tool calls but repeatedly encountered failures when attempting to load the data. Consequently, the entire search tree remained unproductive and failed to reach the modeling phase. This suggests that the system lacked sufficient guidance to advance beyond the initial stages of the pipeline.

\begin{lstlisting}
    {
      "step_number": 22,
      "message_type": "AIMessage",
      "timestamp": "2025-11-20T16:12:41.763597",
      "step_type": "tool_selection",
      "action": "selected_tools_for_execution",
      "tools_selected": [
        "read_data"
      ],
      "tool_calls_detail": [
        {
          "name": "read_data",
          "args": {
            "output": "train_df"
          },
          "id": "call_3did4EVkLGDyw9QRFoiAnUlA",
          "type": "tool_call"
        }
      ],
      "content": ""
    },
    {
      "step_number": 23,
      "message_type": "ToolMessage",
      "timestamp": "2025-11-20T16:12:41.763599",
      "step_type": "tool_result",
      "action": "received_tool_output",
      "content_preview": "Error: TypeError(\"read_data() missing 1 required positional argument: 'filepath'\")\n Please fix your mistakes.",
      "content_length": 109
    },


\end{lstlisting}

\subsubsection{Successful Trajectory}

This trajectory contains 32 tool calls and fits only a CatBoost model. Notably, the submission file produced by this run exhibited a shape mismatch with the hidden test set, resulting in a percentile score of 0. The mismatch arose from errors in the feature engineering stage, which filtered out a portion of the test data.

\begin{lstlisting}
{
      "step_number": 63,
      "message_type": "ToolMessage",
      "timestamp": "2025-11-20T16:13:57.197079",
      "step_type": "tool_result",
      "action": "received_tool_output",
      "content_preview": "Applied label_encode with docstring: \n    Label encode categorical columns.\n    \n    Parameters:\n    -----------\n    df : pd.DataFrame\n        Input DataFrame\n    columns : str or List[str], optional\n        Column(s) to encode. If None, encodes all object/category columns.\n    \n    Returns:\n    ---...",
      "content_length": 701
    },
    {
      "step_number": 64,
      "message_type": "HumanMessage",
      "timestamp": "2025-11-20T16:13:57.197084",
      "step_type": "reward_feedback",
      "action": "generated_reward_feedback",
      "content_preview": "Modeling is still in progress",
      "content_length": 29
    },
    {
      "step_number": 65,
      "message_type": "AIMessage",
      "timestamp": "2025-11-20T16:13:57.197089",
      "step_type": "tool_selection",
      "action": "selected_tools_for_execution",
      "tools_selected": [
        "tune_catboost_classifier"
      ],
      "tool_calls_detail": [
        {
          "name": "tune_catboost_classifier",
          "args": {
            "bindings": {
              "X_train": "X_train",
              "y_train": "Y_train"
            },
            "output": "best_catboost_model"
          },
          "id": "call_WbNtNSt3M3UXFDk5JBj49Nkm",
          "type": "tool_call"
        }
      ],
      "content": ""
    },
    {
      "step_number": 66,
      "message_type": "ToolMessage",
      "timestamp": "2025-11-20T16:13:57.197092",
      "step_type": "tool_result",
      "action": "received_tool_output",
      "content_preview": "Applied tune_catboost_classifier with docstring: \n    Perform hyperparameter tuning for CatBoost Classifier using GridSearchCV.\n\n    Parameters\n    ----------\n    X_train : pd.DataFrame\n        Training features\n    y_train : pd.Series\n        Training target\n    cv : int, default=5\n        Number o...",
      "content_length": 1087
    },
    {
      "step_number": 67,
      "message_type": "HumanMessage",
      "timestamp": "2025-11-20T16:13:57.197097",
      "step_type": "reward_feedback",
      "action": "generated_reward_feedback",
      "content_preview": "Human Feedback: Verified that the modeling was successful",
      "content_length": 57
    },
    {
      "step_number": 68,
      "message_type": "AIMessage",
      "timestamp": "2025-11-20T16:13:57.197102",
      "step_type": "tool_selection",
      "action": "selected_tools_for_execution",
      "tools_selected": [
        "evaluate_classification_model"
      ],
      "tool_calls_detail": [
        {
          "name": "evaluate_classification_model",
          "args": {
            "bindings": {
              "model": "best_estimator",
              "X_test": "X_train",
              "y_test": "Y_train"
            },
            "func_kwargs": {
              "model_name": "CatBoostClassifier",
              "eval_data_label": "train"
            }
          },
          "id": "call_hKlPSrF4ZrPiof5KJTUGpsOb",
          "type": "tool_call"
        }
      ],
      "content": ""
    },
    {
      "step_number": 69,
      "message_type": "ToolMessage",
      "timestamp": "2025-11-20T16:13:57.197105",
      "step_type": "tool_result",
      "action": "received_tool_output",
      "content_preview": "Applied evaluate_classification_model with docstring: \n    Evaluate a trained classification model on data.\n\n    This function is used to evaluate the performance of a trained classification model on a given dataset.\n\n    The dataset can be the anything, such as the test set or the train set.\n\n    U...",
      "content_length": 1830
    },
    {
      "step_number": 70,
      "message_type": "HumanMessage",
      "timestamp": "2025-11-20T16:13:57.197112",
      "step_type": "reward_feedback",
      "action": "generated_reward_feedback",
      "content_preview": "Human Feedback: submission DataFrame was NOT created successfully. Please check the signature of the wrapped function if any, and call it with the correct arguments. The tool signature is This tool reads arguments from the scratchpad using `bindings`, passes them to the internal function.\n\n        A...",
      "content_length": 1979
    },
    {
\end{lstlisting}

\subsection{MCTS-Shaped}

\subsubsection{Failed Trajectory}

The most promising trajectory consisted of 19 steps and terminated at the stage of splitting the combined data into train and test. This indicates that the search never reached the modeling phase.

\begin{lstlisting}
        {
      "step_number": 49,
      "message_type": "AIMessage",
      "timestamp": "2025-11-20T17:41:18.892989",
      "step_type": "tool_selection",
      "action": "selected_tools_for_execution",
      "tools_selected": [
        "drop_feature"
      ],
      "tool_calls_detail": [
        {
          "name": "drop_feature",
          "args": {
            "bindings": {
              "df": "combined_df"
            },
            "output": "combined_df",
            "func_kwargs": {
              "column": [
                "PassengerId",
                "Name",
                "Cabin"
              ]
            }
          },
          "id": "call_W8BQW8V16Jch32pPwgSkJXIN",
          "type": "tool_call"
        }
      ],
      "content": ""
    },
    {
      "step_number": 50,
      "message_type": "ToolMessage",
      "timestamp": "2025-11-20T17:41:18.892992",
      "step_type": "tool_result",
      "action": "received_tool_output",
      "content_preview": "Applied drop_feature with docstring: \n    Drop feature(s) from the DataFrame.\n    \n    Parameters:\n    -----------\n    df : pd.DataFrame\n        Input DataFrame\n    column : str or List[str]\n        Name(s) of the column(s) to drop\n    \n    Returns:\n    --------\n    pd.DataFrame\n        DataFrame wi...",
      "content_length": 662
    },
    {
      "step_number": 51,
      "message_type": "HumanMessage",
      "timestamp": "2025-11-20T17:41:18.892997",
      "step_type": "reward_feedback",
      "action": "generated_reward_feedback",
      "content_preview": "Categorical columns found: [{'column': 'HomePlanet', 'unique_count': 3, 'dtype': 'category'}, {'column': 'CryoSleep', 'unique_count': 3, 'dtype': 'category'}, {'column': 'Destination', 'unique_count': 4, 'dtype': 'category'}]. Columns with dtypes that are not numeric/categorical found: []. Please co...",
      "content_length": 525
    },
    {
      "step_number": 52,
      "message_type": "AIMessage",
      "timestamp": "2025-11-20T17:41:18.893002",
      "step_type": "tool_selection",
      "action": "selected_tools_for_execution",
      "tools_selected": [
        "label_encode"
      ],
      "tool_calls_detail": [
        {
          "name": "label_encode",
          "args": {
            "bindings": {
              "df": "combined_df"
            }
          },
          "id": "call_LZuEYLrfsTWPCfWsh5Ungldl",
          "type": "tool_call"
        }
      ],
      "content": ""
    },
    {
      "step_number": 53,
      "message_type": "ToolMessage",
      "timestamp": "2025-11-20T17:41:18.893005",
      "step_type": "tool_result",
      "action": "received_tool_output",
      "content_preview": "Applied label_encode with docstring: \n    Label encode categorical columns.\n    \n    Parameters:\n    -----------\n    df : pd.DataFrame\n        Input DataFrame\n    columns : str or List[str], optional\n        Column(s) to encode. If None, encodes all object/category columns.\n    \n    Returns:\n    ---...",
      "content_length": 709
    },
    {
      "step_number": 54,
      "message_type": "HumanMessage",
      "timestamp": "2025-11-20T17:41:18.893010",
      "step_type": "reward_feedback",
      "action": "generated_reward_feedback",
      "content_preview": "Human Feedback: Verified that the feature engineering is complete and all columns are numeric (int/float)",
      "content_length": 105
    },
    {
      "step_number": 55,
      "message_type": "AIMessage",
      "timestamp": "2025-11-20T17:41:18.893015",
      "step_type": "tool_selection",
      "action": "selected_tools_for_execution",
      "tools_selected": [
        "split_combined_into_train_test"
      ],
      "tool_calls_detail": [
        {
          "name": "split_combined_into_train_test",
          "args": {
            "bindings": {
              "combined": "combined_df"
            },
            "output": "splitted_dfs"
          },
          "id": "call_LsWLDkHhkrEQ8Ja0rrBMQOXh",
          "type": "tool_call"
        }
      ],
      "content": ""
    },
    {
      "step_number": 56,
      "message_type": "ToolMessage",
      "timestamp": "2025-11-20T17:41:18.893018",
      "step_type": "tool_result",
      "action": "received_tool_output",
      "content_preview": "Applied split_combined_into_train_test with docstring: \n    Split combined data back into train and test using tracking columns.\n    \n    Parameters:\n    -----------\n    combined : pd.DataFrame\n        Combined DataFrame with tracking columns\n    \n    Returns:\n    --------\n    tuple\n        (train_d...",
      "content_length": 607
    },
    {
      "step_number": 57,
      "message_type": "HumanMessage",
      "timestamp": "2025-11-20T17:41:18.893022",
      "step_type": "reward_feedback",
      "action": "generated_reward_feedback",
      "content_preview": "Human Feedback: Verified that the train and test data were split successfully",
      "content_length": 77
    }
  ],
\end{lstlisting}

\subsubsection{Successful Trajectory}

The trajectory spans 23 steps and fits exclusively a LightGBM classifier in the modeling stage. When evaluated on our test split, it attains a leaderboard percentile of 33.11.

\begin{lstlisting}
    {
      "step_number": 61,
      "message_type": "AIMessage",
      "timestamp": "2025-11-20T17:45:34.047940",
      "step_type": "tool_selection",
      "action": "selected_tools_for_execution",
      "tools_selected": [
        "tune_lightgbm_classifier"
      ],
      "tool_calls_detail": [
        {
          "name": "tune_lightgbm_classifier",
          "args": {
            "bindings": {
              "X_train": "X_train",
              "y_train": "Y_train"
            },
            "output": "tuned_model_info"
          },
          "id": "call_LfolkEfImAXxyb5Ha3nNW4cp",
          "type": "tool_call"
        }
      ],
      "content": ""
    },
    {
      "step_number": 62,
      "message_type": "ToolMessage",
      "timestamp": "2025-11-20T17:45:34.047944",
      "step_type": "tool_result",
      "action": "received_tool_output",
      "content_preview": "Applied tune_lightgbm_classifier with docstring: \n    Perform hyperparameter tuning for LightGBM Classifier using GridSearchCV.\n\n    Parameters\n    ----------\n    X_train : pd.DataFrame\n        Training features\n    y_train : pd.Series\n        Training target\n    cv : int, default=5\n        Number o...",
      "content_length": 1092
    },
    {
      "step_number": 63,
      "message_type": "HumanMessage",
      "timestamp": "2025-11-20T17:45:34.047949",
      "step_type": "reward_feedback",
      "action": "generated_reward_feedback",
      "content_preview": "Human Feedback: Verified that the modeling was successful",
      "content_length": 57
    },
    {
      "step_number": 64,
      "message_type": "AIMessage",
      "timestamp": "2025-11-20T17:45:34.047955",
      "step_type": "tool_selection",
      "action": "selected_tools_for_execution",
      "tools_selected": [
        "evaluate_classification_model"
      ],
      "tool_calls_detail": [
        {
          "name": "evaluate_classification_model",
          "args": {
            "bindings": {
              "model": "best_estimator",
              "X_test": "X_train",
              "y_test": "Y_train"
            },
            "func_kwargs": {
              "model_name": "LightGBM",
              "eval_data_label": "train"
            }
          },
          "id": "call_iLtqzfkOBwtWmROl9WDHKO75",
          "type": "tool_call"
        }
      ],
      "content": ""
    },
    {
      "step_number": 65,
      "message_type": "ToolMessage",
      "timestamp": "2025-11-20T17:45:34.047959",
      "step_type": "tool_result",
      "action": "received_tool_output",
      "content_preview": "Applied evaluate_classification_model with docstring: \n    Evaluate a trained classification model on data.\n\n    This function is used to evaluate the performance of a trained classification model on a given dataset.\n\n    The dataset can be the anything, such as the test set or the train set.\n\n    U...",
      "content_length": 1820
    },
    {
      "step_number": 66,
      "message_type": "HumanMessage",
      "timestamp": "2025-11-20T17:45:34.047965",
      "step_type": "reward_feedback",
      "action": "generated_reward_feedback",
      "content_preview": "Human Feedback: submission DataFrame was NOT created successfully. Please check the signature of the wrapped function if any, and call it with the correct arguments. The tool signature is This tool reads arguments from the scratchpad using `bindings`, passes them to the internal function.\n\n        A...",
      "content_length": 1979
    },
    {
      "step_number": 67,
      "message_type": "AIMessage",
      "timestamp": "2025-11-20T17:45:34.047972",
      "step_type": "tool_selection",
      "action": "selected_tools_for_execution",
      "tools_selected": [
        "predict_target"
      ],
      "tool_calls_detail": [
        {
          "name": "predict_target",
          "args": {
            "bindings": {
              "model": "best_estimator",
              "X_data": "X_test"
            },
            "output": "test_predictions",
            "func_kwargs": {
              "model_name": "LightGBM"
            }
          },
          "id": "call_lZQQw4OoT0LANXYCUvjXh1Bp",
          "type": "tool_call"
        }
      ],
      "content": ""
    },
    {
      "step_number": 68,
      "message_type": "ToolMessage",
      "timestamp": "2025-11-20T17:45:34.047975",
      "step_type": "tool_result",
      "action": "received_tool_output",
      "content_preview": "Applied predict_target with docstring: \n    Make predictions using a trained model.\n    \n    Parameters\n    ----------\n    model : Any\n        Trained model (regression or classification)\n    X_data : pd.DataFrame\n        Features to make predictions on\n    model_name : str, default=\"model\"\n        ...",
      "content_length": 1018
    },
    {
      "step_number": 69,
      "message_type": "HumanMessage",
      "timestamp": "2025-11-20T17:45:34.047981",
      "step_type": "reward_feedback",
      "action": "generated_reward_feedback",
      "content_preview": "Human Feedback: Verified that the submission DataFrame was created successfully with 1 columns (boolean) and no missing values.",
      "content_length": 127
    }
  ],


\end{lstlisting}

\subsection{Hierarchical MCTS}

\subsubsection{Failed Trajectory}

This trajectory consists of 55 tool calls and fails at the train data to features and target stage, prior to modeling. Nothing of note in this trajectory, and hence is not reported.

\subsubsection{Successful Trajectory}

The trajectory spans 80 steps and evaluates multiple modeling approaches, attaining a leaderboard percentile of 62.65. Hierarchical MCTS demonstrates the ability to execute substantially longer trajectories, owing to its use of subtask decomposition and task masking.

\begin{lstlisting}
    {
          "step_type": "tool_selection",
          "action": "selected_tools_for_execution",
          "tools_selected": [
            "tune_lightgbm_classifier"
          ],
          "tool_calls_detail": [
            {
              "name": "tune_lightgbm_classifier",
              "args": {
                "bindings": {
                  "X_train": "X_train",
                  "y_train": "Y_train"
                },
                "output": "tuned_lgbm_classifier"
              },
              "id": "call_fM1ezojD8q9aj9f052GtppXm",
              "type": "tool_call"
            }
          ],
          "content": ""
        },
        {
          "step_type": "tool_result",
          "action": "received_tool_output",
          "content_preview": "Applied tune_lightgbm_classifier with docstring: \n    Perform hyperparameter tuning for LightGBM Classifier using GridSearchCV.\n\n    Parameters\n    ----------\n    X_train : pd.DataFrame\n        Training features\n    y_train : pd.Series\n        Training target\n    cv : int, default=5\n        Number o...",
          "content_length": 1092
        },
        {
          "step_type": "reward_feedback",
          "action": "generated_reward_feedback",
          "content_preview": "Human Feedback: Verified that the modeling was successful",
          "content_length": 57
        },
{
          "step_type": "tool_selection",
          "action": "selected_tools_for_execution",
          "tools_selected": [
            "tune_random_forest_classifier"
          ],
          "tool_calls_detail": [
            {
              "name": "tune_random_forest_classifier",
              "args": {
                "bindings": {
                  "X_train": "X_train",
                  "y_train": "Y_train"
                },
                "func_kwargs": {
                  "cv": 5
                },
                "output": "rf_tuned_model_output"
              },
              "id": "call_3OsXBYeYfbCnehW6uvlfA99X",
              "type": "tool_call"
            }
          ],
          "content": "I cannot evaluate the model on the test data because the true target labels are not available for the test set. Instead, I will generate predictions for the test data using the best Random Forest classifier model and save the predictions to the required submission file path."
        },
        {
          "step_type": "tool_result",
          "action": "received_tool_output",
          "content_preview": "Applied tune_random_forest_classifier with docstring: \n    Perform hyperparameter tuning for Random Forest Classifier using GridSearchCV.\n\n    Parameters\n    ----------\n    X_train : pd.DataFrame\n        Training features\n    y_train : pd.Series\n        Training target\n    cv : int, default=5\n      ...",
          "content_length": 1134
        },
        {
          "step_type": "reward_feedback",
          "action": "generated_reward_feedback",
          "content_preview": "Human Feedback: Verified that the modeling was successful",
          "content_length": 57

\end{lstlisting}

\end{document}